\documentclass[12pt,english]{article}
\usepackage{graphicx}
\PassOptionsToPackage{}{amsmath}
\usepackage{float}
\usepackage[T1]{fontenc}
\usepackage[letterpaper]{geometry}
\usepackage{babel}
\usepackage{amsmath}
\usepackage{amsthm}
\usepackage{amssymb}
\usepackage{setspace}
\usepackage{ragged2e}
\usepackage{esint}
\usepackage{array}
\usepackage{adjustbox}
\usepackage{scrextend}
\usepackage{pdfpages}
\usepackage{chngpage}
\usepackage{pdflscape}
\usepackage{everypage}
\usepackage{lipsum}
\usepackage{longtable}
\usepackage{caption}
\usepackage{pgfplots}
\usepackage[utf8]{inputenc}
\usepackage{pifont}
\usepackage[colorinlistoftodos]{todonotes}
\usepackage{graphicx}
\usepackage{siunitx}
\usepackage{appendix}
\usepackage{placeins}
\usepackage{comment}
\usepackage{rotating}
\usepackage{adjustbox}
\usepackage{booktabs}
\usepackage{threeparttable}
\usepackage{makecell} % add this in your preamble7
\usepackage{siunitx} % for thousands separator and alignment
\newcommand{\Lpagenumber}{\ifdim\textwidth=\linewidth\else\bgroup
  \dimendef\margin=0 %use \margin instead of \dimen0
  \ifodd\value{page}\margin=\oddsidemargin
  \else\margin=\evensidemargin
  \fi
  \raisebox{\dimexpr -\topmargin-\headheight-\headsep-0.5\linewidth}[0pt][0pt]{%
    \rlap{\hspace{\dimexpr \margin+\textheight+\footskip}%
    \llap{\rotatebox{90}{\thepage}}}}%
\egroup\fi}
\AddEverypageHook{\Lpagenumber}%

\usepackage{threeparttable}
\newcolumntype{P}[1]{>{\centering\arraybackslash}p{#1}}
\usepackage[unicode=true,pdfusetitle,
 bookmarks=true,bookmarksnumbered=false,bookmarksopen=false,
 breaklinks=false,pdfborder={0 0 0},pdfborderstyle={},backref=false,colorlinks=True,linkcolor=blue,citecolor=blue]
 {hyperref}
\graphicspath{ {.} }
\usepackage{xcolor} % allow colors

\usepackage[bottom,flushmargin]{footmisc} % keep footnote at the bottom of page.
\usepackage{ragged2e}
\usepackage{textgreek}
\usepackage{multirow, makecell}
\usepackage{fancyhdr}
\usepackage{tabularx}
\usepackage{makecell}
\usepackage{bookmark}
\hypersetup{colorlinks,
linkcolor=blue,
urlcolor=blue}
\def\fillandplacepagenumber{%
 \par\pagestyle{empty}%
 \vbox to 0pt{\vss}\vfill
 \vbox to -1cm{\baselineskip0pt
   \hbox to\linewidth{\hss}%
   \baselineskip\footskip
   \hbox to\linewidth{%
     \hfil\thepage\hfil}\vss}}
\usepackage[authoryear,round,sort]{natbib}
\usepackage{booktabs} 

\makeatletter
\usepackage{float}
\usepackage{subfloat}
\usepackage{lmodern}
\usepackage{setspace}
\usepackage{amsfonts}
\usepackage{booktabs}
\usepackage{tablefootnote}
\usepackage{comment}
\usepackage{caption}

\usepackage{subcaption}

\usepackage{float}

\usepackage{babel}\usepackage{babel}
\usepackage{pdflscape}

\renewcommand{\@seccntformat}[1]{{\csname the#1\endcsname}.\hspace{0.5em}}
\renewcommand{\section}{\@startsection {section}{1}{0pt}{3.5ex}{0.5ex}{\bf\large}}
\renewcommand{\subsection}{\@startsection {subsection}{1}{0pt}{3.5ex}{0.5ex}{\it\large}}
\renewcommand{\subsubsection}{\@startsection {subsubsection}{1}{0pt}{3.5ex}{2.3ex}{\it\large}}

\renewcommand{\thefootnote}{\fnsymbol{footnote}}

\begin{document}

\title{}
\author{}
\date{}
\maketitle
\thispagestyle{empty}

\vspace{-1in}
\begin{center}
\Large{\textbf{The Role of AI in Online Reviews}}
\end{center}

\vspace*{0.1in}

\begin{center}
    
\renewcommand{\thefootnote}{\fnsymbol{footnote}}

Valeria Lerman$^{a,*}$, Oren Rigbi$^{a}$, and Yaniv Dover$^{a,b,c}$

\begingroup
\renewcommand{\thefootnote}{\alph{footnote}}

\footnotetext[1]{The Hebrew University Business School, The Hebrew University of Jerusalem, Jerusalem 9190501, Israel}

\footnotetext[2]{Department of Cognitive and Brain Sciences, Faculty of Humanities, The Hebrew University of Jerusalem, Jerusalem 9190501, Israel}

\footnotetext[3]{The Federmann Center for the Study of Rationality, The Hebrew University of Jerusalem, Jerusalem 9190501, Israel}

\endgroup

\begingroup
\renewcommand{\thefootnote}{\fnsymbol{footnote}}

\footnotetext[1]{Corresponding author. Email: valeria.lerman@mail.huji.ac.il}

\endgroup

\vspace{0.2in}

\today 

\end{center}
\setstretch{1}

\begin{center}
~\\ ~\\

\begin{abstract}
The rapid adoption of large language models (LLMs) creates new opportunities for strategic content generation on online platforms, including potentially harmful forms of manipulation that may undermine platform effectiveness and reshape platform dynamics. However, measuring such activity is difficult because AI-generated content is rarely directly observable. We introduce an empirical approach that leverages discrete LLM supply shocks - abrupt changes in model prices and capabilities, and contrasts verified with non-verified reviews to identify changes in platform activity associated with generative AI supply improvements. We apply this approach to more than 13 million reviews from Trustpilot, one of the leading online platforms for business reviews. A robust finding is that following LLM supply shocks, unverified reviews shift toward greater negativity: more 1-stars, fewer 5-stars, and lower ratings, with effects driven primarily by new model releases and concentrated among firms with the lowest and highest review volumes, suggesting that strategic AI use may reshape platform competition dynamics. We further find that LLM supply shocks trigger short, concentrated bursts of review activity. Together, these findings suggest that generative AI is already reshaping how reputation and competition operate on online platforms.
\end{abstract}

\end{center}

\textbf{Keywords:} Generative AI; Large Language Models; Online Reviews; Digital Platforms; User-Generated Content

\setstretch{1}
\pagebreak
%%%%%%%%%%%%%%%%%%%%%%%%%%%%%%%%%%%%%%%%%%%%%%%%%%%%%%%%%%%%%%%%%%%%%%%

\setcounter{page}{1} % sets the current page number
\renewcommand{\thefootnote}{\arabic{footnote}}
\setcounter{footnote}{0}
\onehalfspacing

\section{Introduction}\label{intro}

In recent years, generative artificial intelligence (GenAI) has diffused rapidly across a wide range of domains, transforming both the automation of digital tasks and the production of content. Advances in large language models (LLMs) have enabled AI systems not only to generate coherent, contextually relevant, and human-like text, but also, increasingly, to browse the internet, interact with digital platforms, and perform actions autonomously. These capabilities have fundamentally altered the economics of text production by sharply reducing its marginal cost and enabling content to be generated at unprecedented scale and speed (\cite{brown2020language, bommasani2021opportunities}). As a result, text-intensive activities that traditionally required substantial human effort, including drafting news and opinion articles, producing creative content, generating social media posts, and composing online reviews, can now be partially or fully automated.

Empirical studies indeed show that AI-generated text is often indistinguishable from human-written content, particularly when evaluated by non-expert readers (\cite{clark2021all, kreps2022all}). Because content on digital platforms can significantly shape market dynamics and economic outcomes, this high degree of realism raises important questions about authenticity and its broader implications for information ecosystems. The increasing presence of machine-generated content blurs the boundary between human and automated expression, complicating the ability of businesses, users, and platforms to assess the origin and credibility of online information.

One prominent narrative reflecting public concern is the so-called “dead internet theory,” which suggests that a substantial portion of online content is already generated by automated systems rather than humans (\cite{baronio2025deadinternet, down2025aislop, murray2025deadinternet,levy2026aislop}). Although this claim is sometimes raised in public debate, it lacks rigorous empirical support. Nevertheless, it reflects broader concerns about the growing role of artificial agents in digital environments. Even if the internet is far from “dead,” there is increasing evidence that automated content generation is becoming a growing component of online activity (\cite{ferrara2016rise, muzumdar2025dead, walter2025artificial}).

The implications of this shift are particularly salient in the context of user-generated content (UGC), which plays a central role in modern digital economies. Social media platforms, for example, are highly dependent on user contributions that influence public opinion, consumer behavior, and even political outcomes (\cite{allcott2017social}). Integrating AI-generated content into these ecosystems reduces dissemination costs, but increases the risk of low-quality, misleading, or strategically manipulated content. Similarly, informational platforms such as Wikipedia or question-and-answer forums depend on crowd-sourced knowledge production. Integration of AI tools may improve productivity, but could also affect content reliability and editorial norms (\cite{shin2026ai, park2024rise, marcellino2023rise}). 

The issue becomes even more pressing in the context of commercial content. Online reviews, product descriptions, and marketing materials are essential inputs in consumer decision-making processes (\cite{rachmiani2024impact, lackermair2013importance}) and were repeatedly shown to affect important economic outcomes (\cite{pocchiari2025online, alzate2021online, huang2020impact}). The incentives for businesses to generate such content at scale using AI without disclosing its origin are clear. Although such practices are unethical, they offer an efficient way to shape their own and others’ online reputations, increase demand, and achieve other desirable outcomes. If firms engage in the production of hard-to-identify AI-generated online reviews, there is a considerable risk to market authenticity and transparency, and consequently the usefulness of online reviews in digital markets. Moreover, Generative AI may alter competitive dynamics on commercial platforms in ways that further undermine the existing digital ecosystem. These concerns warrant systematic scientific investigation.

Here, we focus on online review platforms, which are central to the digital economy and play a key role in shaping consumer decisions and market outcomes, including demand, pricing, and firm reputation (\cite{qiu2023online, burton2024reviews}). However, their credibility may be undermined by fake and manipulated reviews, which firms may strategically use to increase ratings, improve brand image, or harm competitors (\cite{mayzlin2014promotional, lim2025rise, martinez2021fake, he2022market}). These distort consumer signals and can produce inefficient market outcomes.

Generative AI may significantly worsen these challenges by making it far cheaper and easier to produce large volumes of human-like fake reviews. Compared to earlier fake reviews, AI-generated content has the potential to be more coherent, diverse, and contextually relevant, increasing both the scale of potential manipulation and the difficulty of detection (\cite{inbook, gupta2024recent, zhao2025ai, meng2025large, knight2023generative}).

Some recent studies have attempted to distinguish AI-generated online reviews from authentic human-written reviews by identifying differences in their linguistic characteristics. These studies find that AI-generated reviews are often more readable and coherent but tend to be less specific, emotional, and empathetic, and tend to follow a more mechanical structure (\cite{zhao2025ai}). However, these findings may not generalize across contexts and settings or persist over time, given the rapid and substantial advances in AI capabilities. The findings also suggest that AI-generated deception differs from traditional fake reviews and can evade detection methods that rely on psychological cues associated with human behavior. Related research finds that in their case AI-generated reviews are often higher-rated, posted by users with limited platform history, and are more readable but less linguistically complex than authentic reviews and appear more common among lower-traffic businesses (\cite{gambetti2023dissecting}). Despite these studies, there is still no conclusive evidence that AI-generated reviews can be systematically identified using textual markers. On the contrary, other studies show that both humans and advanced language models have a hard time detecting AI-generated reviews (\cite{meng2025large, santos2025improving, agrahari2025can}).
Another strand of research attempts to develop more advanced detection methods, combining textual features with statistical signals such as outlier patterns in review distributions. Although these approaches seem to somewhat improve performance, they still depend on model-specific assumptions and imperfect ground-truth proxies, underscoring the ongoing difficulty in identifying AI-generated reviews at scale (\cite{luo2026ai, gambetti2023combat}).

Taken together, the literature is mixed. AI-generated reviews may differ from human-written ones, but evidence on their detectability and prevalence is inconsistent, in part because studies rely on different data, labels, and evaluation settings, often using synthetic samples. This highlights the need to use novel approaches to better understand how firms use Generative AI tools in practice — whether for self-promotion, competitive manipulation, or both—and with what consequences for market outcomes.

To examine the real-world use of generative AI for online review manipulation, we develop a novel empirical strategy that leverages rapidly changing exogenous variation in the cost and capabilities of AI-generated text. Specifically, we use data from Trustpilot, a large online platform hosting consumer reviews of firms and services for the period of 2023 - 2024, and leverage OpenAI's discrete API price reductions and the introduction of more affordable and more efficient model tiers as means of supply-side shifts. This abrupt reduction in the cost and improvement in the capabilities of generative AI provide a unique opportunity to study whether and how firms respond to changes in the economic feasibility of using generative AI for content production. Both abrupt price reductions and new model launches provide plausibly exogenous shocks, as their precise timing is generally unknown in advance and is unlikely to be driven by the activities of Generative AI producers on Trustpilot.\footnote{Although price reductions or model launches may occasionally be anticipated by a day or two, such limited advance notice is unlikely to threaten identification. Bias would arise only if actors immediately changed activity before the official event and before the lower price and new capabilities are introduced, which does not seem likely in our context.} We focus on OpenAI because it was the leading provider in the LLM API market during that period \cite{tully2024state, wang2024generative}. In what follows, we use the term ‘LLM supply shocks’ to jointly refer to these API price reductions and the introduction of more affordable or efficient model tiers.

Our approach leverages the exogeneity of unanticipated discrete price reductions and new language model introductions along with the verified-reviews feature of the Trustpilot platform. In particular, we exploit the distinction between verified and unverified reviews on the platform, in the spirit of \cite{mayzlin2014promotional} and \cite{luca2016fake}. Verified reviews are linked to confirmed transactions or experiences and are therefore more likely to reflect genuine consumer activity. Verified reviews form part of the substantial efforts that Trustpilot reports undertaking to detect and remove fake reviews (\cite{trustpilot_trust}).\footnote{The platform also reports having flagged roughly 6\% of submitted unverified reviews as fake in recent years, indicating an active commitment to identifying and removing fraudulent content, including potentially AI-generated reviews. Thus, one implication is that any AI-generated activity detected in our analysis reflects activity that remains observable despite the platform’s filtering efforts. We discuss the implications of this moderation process below.} In contrast, unverified reviews face fewer credibility constraints and are more susceptible to strategic manipulation, including the potential use of AI-generated content. Under this assumption, unverified reviews are more likely to respond to changes in the cost and capabilities of generative AI.

By exploiting variation along two dimensions — immediate time (shortly before vs. after LLM supply shocks) and review type (verified vs. unverified) — we implement a difference-in-differences framework that compares changes over time in review characteristics across groups. We focus on short time windows around the LLM supply shocks to isolate the effect of each of these events and reduce contamination from longer-term effects. This design allows us to isolate the differential impact of changes in the cost and capabilities of generative AI on reviews that are more likely to incorporate generative AI. Our specifications include time and company fixed effects, so identification comes from within-company changes around each LLM supply shock, and because verified reviews serve as a control group, the estimates absorb broader common shocks that affect both verified and unverified reviews, increasing confidence that any remaining effect reflects AI-driven changes in unverified review activity.
 
Our empirical analysis yields two central sets of findings. First, we document systematic changes in the distribution of ratings following exogenous changes in the cost and capabilities of generative AI, driven by OpenAI price reductions and new model releases. Across specifications, unverified reviews experience a statistically significant decline in average ratings relative to verified reviews in the seven days following the LLM supply shock. This shift is driven by an increase in the share of one-star reviews and a corresponding decrease in the share of five-star reviews. Although the estimated effects are modest in percentage-point terms, they represent a lower bound on GenAI activity in online reviews. Our approach captures only marginal responses to changes in model prices and capabilities, and we do not observe reviews that were flagged and removed by the platform. We therefore cannot estimate the absolute prevalence of AI-generated activity across the platform. Given these limitations, the observed effects are nevertheless economically meaningful at Trustpilot’s scale and imply substantial shifts in the overall distribution of ratings. Furthermore, we find that these effects are driven by the release of newer, more capable language models rather than by price reductions, and are strongest for reviews of the largest and smallest firms, as proxied by platform-level activity.

These results suggest that lower-cost and more efficient generative AI is associated with an increase in negative review activity rather than positive self-promotion. A plausible interpretation is that firms use AI-generated content strategically to target competitors rather than primarily to promote themselves. The greater capabilities and efficiency of newer models may enable more sophisticated competitive strategies, while negative reviews may also carry a lower risk of detection than attempts to artificially inflate a firm’s own ratings, particularly given the platform’s intensive efforts to identify fraudulent activity.

In addition, although data limitations prevent us from detecting an increase in aggregate review volume following price reductions, we do find a significant short-term surge of review activity after the introduction of newer, more capable models. This pattern suggests that AI tools are used not merely to generate reviews, but as part of a specific, concentrated review-generation tactic.

Taken together, these findings contribute to the growing literature on the economic and behavioral implications of generative AI by providing causal evidence of how improvements in AI capabilities and reductions in AI usage costs shape online information environments. More specifically, our results provide some of the first empirical evidence that AI-generated reviews are actively used in practice. They further suggest that, at least on Trustpilot, such reviews are deployed primarily as a competitive strategy to harm rival firms rather than as a tool for enhancing firms' own reputations. This interpretation should, however, be tempered by the possibility that the platform’s aggressive filtering policies more effectively remove self-promotional fake reviews, rendering them less observable even if they are present. From a practical perspective, the findings raise important concerns for digital platforms and regulators, suggesting that cheaper and more accessible AI tools may intensify challenges related to review authenticity, platform trust, platform dynamics, and consumer welfare.

% \subsection{Structure of the Paper}

The remainder of the paper is organized as follows: Section \ref{Sec:DestStat} describes the data sources, data preparation procedures, and descriptive statistics. Section \ref{sec:methods} presents the empirical methodology. Section \ref{sec:results} reports the effects of LLM supply shocks on review ratings and review-volume categories. Section \ref{sec:robust} presents robustness checks and additional heterogeneity analyses. Finally, Section \ref{sec:conclusion} concludes.

\section{Data and Descriptive Statistics}\label{Sec:DestStat}
This study combines data from two primary sources. First, we collect reviews from the Trustpilot platform, which provides reviews of different services, both online and offline. Second, we construct a dataset of OpenAI supply shock dates. By combining these datasets, we are able to examine how changes in the cost and capabilities of generative AI affect online reviews.

\subsection{Data Sources}\label{subsec:data sources}

\subsubsection{Trustpilot Reviews Data}\label{subsubsec:Trustp Data}

The primary dataset used in our analysis consists of 13,818,281 reviews collected from the Trustpilot platform, which hosts public reviews of services.\footnote{The data were obtained through Bright Data, a third-party data provider, and are not publicly accessible.} From this dataset we utilize several key variables for our analysis: review creation date, review rating (between 1 and 5), review verification status, and company identifier. The specific ways in which these variables are used in the empirical analysis are described in Section \ref{sec:methods}. We perform the analysis using reviews written between January 1, 2023, and December 31, 2024 (inclusive). It is also important to note that the Trustpilot data has already undergone platform-level moderation and cleaning. \footnote{According to Trustpilot’s transparency reports, the removal of fake reviews remains relatively stable year-over-year at approximately 6\%, with the majority of such reviews (about 82\%) identified through Trustpilot’s automated detection technology (\cite{trustpilot_transparency}). Trustpilot attributes this capability to continuous investments in advanced detection systems, including machine learning and AI-based models that leverage the platform’s growing volume of review data. These systems analyze hundreds of data points and identify suspicious behavioral patterns and anomalies (e.g., unusual reviewing activity or coordinated behavior) to detect guideline violations.}

\subsubsection{OpenAI LLM Supply Shock Dates}\label{subsub:price shock dates}

To construct the dataset of LLM supply shock events, we manually collected the dates on which OpenAI announced price reductions and/or released new models. We focus on OpenAI-related events because it was the dominant actor in the generative AI market during the 2023--2024 period, both in terms of technological leadership and widespread adoption (\cite{enterpriseappstoday_openai_stats, firstpagesage_market_share, softwareoasis_chatgpt_dominance, feedough_openai_stats, axios_chatgpt_users}). The release of ChatGPT and subsequent model iterations (e.g., GPT-4 and GPT-4o) drove rapid diffusion of AI technologies across industries, making OpenAI the primary source of market-wide pricing shocks (\cite{raman2024exploring, zhang2024exploring}). As a result, price changes by OpenAI are likely to capture the most economically meaningful variation in AI costs during our sample period, whereas competing providers had more limited adoption or entered the market later. Since we could not locate any single centralized source that lists all historical supply shocks, we compiled this information from multiple publicly available sources that we gathered after a comprehensive search. The primary sources include official announcements and updates published on OpenAI’s website, as well as reports from artificial intelligence–focused technology blogs (\cite{openai_forum_cost_comparison, voicebot_openai_price_reduction_2023, openai_devday_2023, openai_embedding_api_updates_2024, openai_hello_gpt4o_2024, openai_gpt4o_mini_2024, spiceworks_openai_structured_outputs_2024, towardsagi_o1_pricing_2024, openai_devs_twitter_pricing, openai_realtime_api_updates_2024, verge_chatgpt_api_2023}). To ensure comprehensive coverage, we conducted multiple systematic searches across OpenAI’s official communications (including blog posts, release notes, and developer updates) as well as leading AI-focused technology outlets. We cross-validated information across sources and compared overlapping reports to minimize omissions and ensure consistency in the documented pricing changes.

\begin{figure}[H]
    \centering

    \captionsetup{
        width=0.8\textwidth,
        justification=centering
    }
    \caption{OpenAI Model Pricing Reductions and Releases, 2023--2024 (normalized to initial price = 100\%)}
    \label{fig:price_percent}

    \begin{minipage}{0.8\textwidth}
        \centering
        \includegraphics[width=\textwidth]{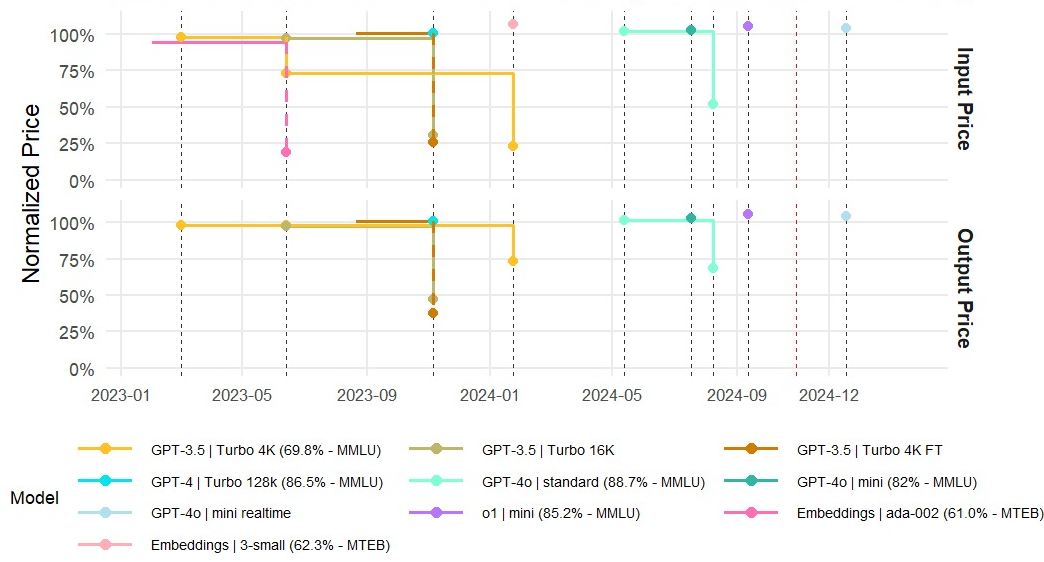}

        \vspace{0.3em}

        {\scriptsize
\justifying
\noindent\textit{Notes:} This figure is based on the pricing data reported in Table~\ref{tab:openai_price_reductions} and on the raw (non-normalized) price series presented in Figure~\ref{fig:price_raw}. Prices are normalized to 100\% at the initial observed price for each model. The horizontal axis displays calendar dates over the period 2023--2024, while the vertical axis reports normalized prices (in percentage terms). The figure consists of two panels: the upper panel shows input prices and the lower panel shows output prices. Colors correspond to different models, with similar color schemes used for models belonging to the same family. The values reported in parentheses in the legend denote benchmark performance scores for each model. Specifically, they correspond to Massive Multitask Language Understanding (MMLU) scores for language models and Massive Text Embedding Benchmark (MTEB) scores for embedding models. These benchmark scores are included only to provide an approximate indication of each model's capabilities. Vertical dashed lines indicate the dates of the LLM supply shocks used in the empirical analysis and listed in Table~\ref{tab:openai_price_reductions}. The red dashed vertical line denotes the LLM supply shock associated with the \textit{gpt-realtime} model; since detailed pricing information is unavailable, this event is indicated without corresponding price series. Solid circular markers denote the pricing observations included in the analysis, namely prices corresponding to price reductions or to the release of new, more efficient models. For some models, such as GPT-3.5 Turbo FT and text-embedding-ada-002, the initial release price is shown in the series but is not marked with a solid point because it does not correspond to a price reduction or a lower-priced model release and was therefore not included in the analysis. In addition, the release of text-embedding-ada-002 predates the sample period and is not displayed, as the figure is restricted to dates from January~1,~2023 onward. Embedding models do not report output prices, as they generate vector representations rather than textual outputs. Finally, slight horizontal jitter is applied to improve visual separation between overlapping series.
\par}

    \end{minipage}

\end{figure}

Our working assumption is that these sources report LLM supply shocks simultaneously or shortly after they are announced, as such changes are typically communicated publicly and quickly disseminated within the AI community. Using these sources, we identified the relevant timings of LLM supply shocks, and constructed a timeline of events that we use as exogenous shocks in our empirical analysis. Table~\ref{tab:openai_price_reductions} summarizes the dates of LLM supply shocks included in the analysis, along with the corresponding posted input and output prices for the affected models, while Figure~\ref{fig:price_percent} illustrates the evolution of these prices over time after normalizing each model's initial observed price to 100\%.

\begin{table}[H]
\centering

\caption{OpenAI Model Pricing Reductions and Releases}
\label{tab:openai_price_reductions}

\begin{threeparttable}

\begin{adjustbox}{width=\textwidth}
\begin{tabular}{llllll}
\toprule
Date & Model &
\makecell{Input price \\ (1000 tokens, \$)} &
\makecell{Output price \\ (1000 tokens, \$)} &
Model type & Event type \\
\midrule

2023-03-01 & ChatGPT 3.5 Turbo 4K & 0.002 & 0.002 & Language model & Price reduction \\
\specialrule{0.2pt}{0pt}{0pt}

2023-06-13 & ChatGPT 3.5 Turbo 4K & 0.0015 & 0.002 & Language model & Price reduction \\
 & ChatGPT 3.5 Turbo 16K & 0.003 & 0.004 & Language model & Price reduction \\
 & text-embedding-ada-002\tnote{a} & 0.0001 & N/A\tnote{b} & Embedding model & Price reduction \\
\specialrule{0.2pt}{0pt}{0pt}

2023-11-06 & ChatGPT 3.5 Turbo 16K & 0.001 & 0.002 & Language model & Price reduction \\
 & ChatGPT 3.5 Turbo 4K FT & 0.003 & 0.006 & Language model & Price reduction \\
 & GPT-4-Turbo 128K & 0.01 & 0.03 & Language model & New model release \\
\specialrule{0.2pt}{0pt}{0pt}

2024-01-24 & ChatGPT 3.5 Turbo 4K & 0.0005 & 0.0015 & Language model & Price reduction \\
 & text-embedding-3-small\tnote{a} & 0.00002 & N/A\tnote{b} & Embedding model & New model release \\
\specialrule{0.2pt}{0pt}{0pt}

2024-05-13 & GPT-4o & 0.005 & 0.015 & Language model & New model release \\
\specialrule{0.2pt}{0pt}{0pt}

2024-07-18 & GPT-4o mini & 0.00015 & 0.0006 & Language model & New model release \\
\specialrule{0.2pt}{0pt}{0pt}

2024-08-09 & GPT-4o & 0.0025 & 0.01 & Language model & Price reduction \\
\specialrule{0.2pt}{0pt}{0pt}

2024-09-12 & o1-mini & 0.003 & 0.012 & Language model & New model release \\
\specialrule{0.2pt}{0pt}{0pt}

2024-10-30 & gpt-realtime & N/A\tnote{b} & N/A\tnote{b} & Realtime model & Price reduction \\
\specialrule{0.2pt}{0pt}{0pt}

2024-12-18 & GPT-4o mini realtime (text) & 0.0006 & 0.0024 & Realtime model & New model release \\

\bottomrule
\end{tabular}
\end{adjustbox}

\begin{tablenotes}[flushleft]
\footnotesize

\item
\parbox{\textwidth}{%
\textit{Notes:} The table summarizes key OpenAI API pricing reductions and model introductions over 2023--2024. For each event, we report the date, model, input and output token prices, model type, and whether the event corresponds to a price reduction or a new model release. These events form the basis for the LLM supply shocks used in the empirical analysis.
}

\vspace{0.3em}

\item[a]
\parbox{\textwidth}{%
Embedding models process input text into vector representations and therefore do not generate text output. As a result, they have an input price but no corresponding output price.
}

\vspace{0.3em}

\item[b]
\parbox{\textwidth}{%
For this event, the price reduction was implemented through prompt caching rather than a change in the posted base token price. Cached text inputs were discounted by 50\%, and cached audio inputs were discounted by 80\%.
}

\end{tablenotes}

\end{threeparttable}
\end{table}

In order to conduct the empirical analysis, we use the Trustpilot platform review data to construct the online reviews activity around the OpenAI supply shocks. The detailed procedures used to prepare the dataset are described in the following section.

\subsection{Data Preparation}
\label{sec:data_preparation}

\subsubsection{Data Filtration and Aggregation}\label{subsubsec:data filt aggr}

For the purpose of the analyses, we keep only reviews with valid creation dates and valid ratings (i.e., that are between 1 and 5). This process results in the exclusion of 270 reviews from the 2023--2024 sample which includes 13.8M reviews. In addition, our analysis includes only reviews written within the relevant seven days before and seven days after each event's time windows which amounts to about 2.8M reviews. For better interpretability, we excluded the event day itself. We aggregated review-level data at the company–date-verification level. Specifically, observations are grouped by company identifier, review verification status (verified or non-verified), and the date on which the review was written. As a result, each observation in the aggregated dataset corresponds to the set of all reviews posted for a given company on a specific date with a given verification status.

For each company–date–verification status group, we compute several statistics that we will use as the outcome variables that characterize review provision activity. For these statistics, we assume that a company began its activity on the platform when the first of its reviews were recorded in the data. The outcome variables are: the total number of reviews, the average review rating, the count and proportion of reviews in each rating category (from 1 to 5), the average number of words per review, the count of daily review surges (batches of reviews per company, verification status, and day; see Section~\ref{subsubsec: res base volume} for a detailed description of these measures), and a measure of review text homogeneity which captures the degree of the similarity among reviews within a company-date-verification group.

\subsection{Descriptive Statistics}\label{desc stats}

The summary statistics for the main variables in our Trustpilot dataset are reported in Table \ref{tab:summary_stats}. The table presents descriptive statistics for both the full dataset of reviews collected between January 1, 2023, and December 31, 2024, and the event-window sample used in our main analyses. The full dataset contains 13,818,281 observations corresponding to 136,750 companies, whereas the event-window sample, which includes reviews posted during the seven days before and the seven days after each LLM supply shock (excluding the event day itself), contains 2,797,199 observations from 72,298 companies.

\begin{table}[H]
\centering
\caption{Summary Statistics for the Full Sample and the Event-Window Analysis Sample}
\label{tab:summary_stats}

\begin{threeparttable}

\begin{adjustbox}{max width=\textwidth}
\begin{tabular}{lrr}
\toprule
Statistic
& \multicolumn{1}{c}{Full sample}
& \multicolumn{1}{c}{Event-window sample} \\
\midrule

Number of observations
& 13,818,281
& 2,797,199 \\

Number of unique companies
& 136,750
& 72,298 \\

Number of unique review dates
& 731
& 140 \\

Mean review rating
& 4.18
& 4.19 \\

Standard deviation of review rating
& 1.47
& 1.47 \\

Average number of reviews across all companies per day
& 18,903.26
& 19,979.99 \\

Average number of reviews per company
& 101.05
& 38.69 \\

Average review length (words)
& 33.24
& 32.92 \\

Unverified reviews (\%)
& 40.20
& 40.17 \\

Verified reviews (\%)
& 59.80
& 59.83 \\

Rating: 1 (\%)
& 15.10
& 14.97 \\

Rating: 2 (\%)
& 2.34
& 2.29 \\

Rating: 3 (\%)
& 3.43
& 3.44 \\

Rating: 4 (\%)
& 7.58
& 7.77 \\

Rating: 5 (\%)
& 71.54
& 71.53 \\

\bottomrule
\end{tabular}
\end{adjustbox}

\end{threeparttable}
\end{table}

The distribution of review ratings is strongly skewed toward positive evaluations. In the full sample, the mean review rating is 4.18 (SD = 1.47), with 5-star and 1-star reviews accounting for 71.54\% and 15.10\% of all reviews, respectively. Ratings of 2, 3, and 4 stars are considerably less common, representing 2.34\%, 3.43\%, and 7.58\% of the sample. This distribution is consistent with the well-documented polarization of online review ratings reported in the literature (\cite{hu2009online, chevalier2006effect, tanase2024online}). Importantly, the event-window sample used in our analyses exhibits nearly identical characteristics, with very similar rating distributions, average rating, and review length. This suggests that the observations included in the event windows are highly representative of the overall population of reviews during the study period.

Table \ref{tab:summary_stats} also shows that verified reviews account for 59.80\% of the full dataset, while unverified reviews account for the remaining 40.20\%. The corresponding proportions in the event-window sample are virtually identical (59.83\% and 40.17\%, respectively; see Section \ref{sec:methods} for a detailed description of the verification process). This similarity is advantageous for our empirical design, as it indicates that the event-window sample closely resembles the full dataset while allowing verified reviews to serve as a meaningful control group. Finally, reviews contain an average of 33.24 words in the full sample and 32.92 words in the event-window sample,\footnote{The calculation of the average review length is based only on non-empty reviews.} suggesting that reviews on the platform tend to be relatively concise regardless of the sample considered.

\begin{figure}[H]
    \centering

    \caption{Daily Number of Reviews Over Time (2023--2024)}
    \label{fig:reviews_over_time}

    \begin{minipage}{0.9\textwidth}
        \centering
        \includegraphics[width=\textwidth]{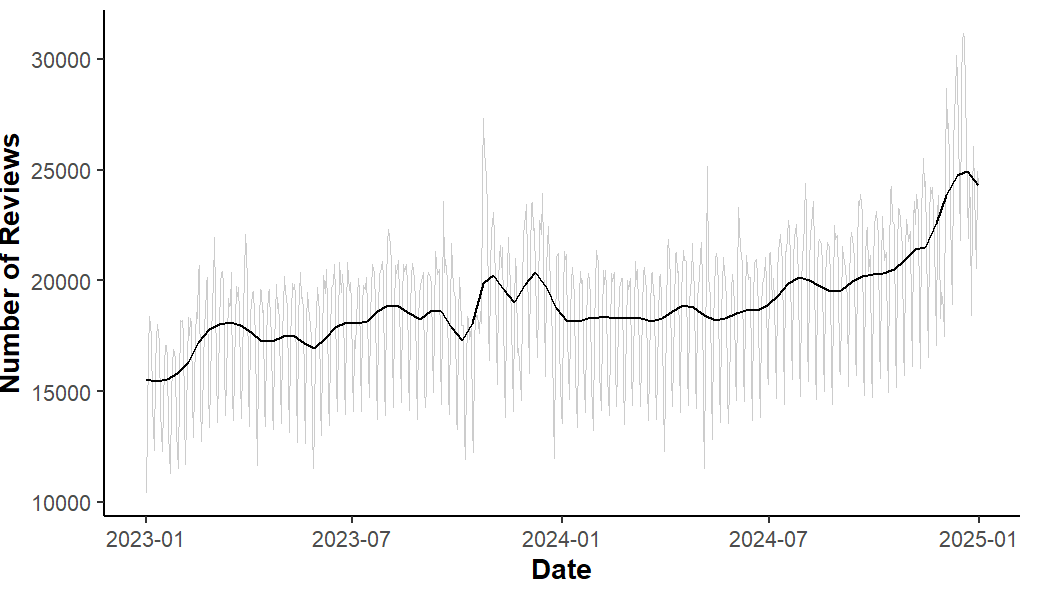}

        \vspace{0.3em}

        {\raggedright
        \scriptsize
        \textit{Notes:} The grey line shows the raw daily number of reviews, while the black line represents the smoothed trend. The figure is based on $N = 13{,}818{,}281$ observations.
        \par}
    \end{minipage}

\end{figure}

Figure \ref{fig:reviews_over_time} illustrates the evolution of review activity over time with a daily resolution during the 2023--2024 period. The gray line represents the daily number of reviews and the black line shows a smoothed trend line using a LOESS procedure. The figure reveals a moderate and relatively stable increase in review activity over the sample period, with no apparent anomalous shifts in platform-wide review volume.

\section{Methodology}\label{sec:methods}

This section outlines the empirical strategy used to estimate the effect of OpenAI’s supply shocks on the characteristics of online reviews. Leveraging temporal variation in AI usage costs and model quality and efficiency, we employ a difference-in-differences design that compares changes in outcome variables for unverified reviews relative to verified reviews before versus after each LLM supply shock.

\subsection{Baseline Difference-in-Differences Analysis}\label{subsec: base did}

To estimate the effect of LLM supply shocks on online reviews, we implement a difference-in-differences (DiD) regression with company and time fixed effects. The objective is to measure how review properties change following changes in the cost and capabilities of OpenAI models. Importantly, the baseline specification treats all LLM supply shocks as equivalent events, regardless of their magnitude or any concurrent changes in model quality or performance. In this approach, we compare changes over time (before-versus-after an event) between verified and unverified reviews within a company. In additional analyses presented in Section \ref{sec:results}, we relax this assumption and examine heterogeneity across different types of events, distinguishing between abrupt price reductions, new lower-priced more-efficient model releases, and specifications that combine both categories of shocks.

Our identification strategy relies on the assumption that unverified reviews are much more likely to be generated or assisted by AI tools, while verified reviews are linked to specific consumers with a confirmed genuine experience with the business who opt to review. Verified reviews are therefore considerably less likely to use AI-generated content created using API (\cite{trustpilot_verified}). Under this assumption, verified reviews serve as a control group, while unverified reviews constitute the treatment group that is likely to respond to changes in AI usage costs.

For each event, we construct an event window that includes the seven days prior to the LLM supply shock and the seven days following the supply shock. The event date itself is excluded from the analysis. We use a relatively narrow seven-day window to limit the influence of potentially confounding events occurring around the LLM supply shocks. Extending the window to longer periods, such as 10 or 14 days, would increase the likelihood that other contemporaneous events unrelated to the focal supply shock affect review activity, making it more difficult to attribute observed changes to the shock itself. In addition, longer windows would result in overlap between the event windows of some LLM supply shocks in our sample, potentially confounding the estimated effects of distinct shocks. The seven-day window therefore provides a balance between allowing sufficient time to capture changes following each shock and maintaining a sufficiently narrow window to isolate its effect. The analysis uses the aggregated outcome and predictor variables for the before and after periods.
Specifically, we estimate the following difference-in-differences specification:

\begin{equation}
\label{eq:did}
Y_{cvt} = \alpha_c + \gamma_{w(t)} + \beta_1 Treatment_v + \beta_2 After_t + \beta_3 (After_t\times Treatment_v) + \varepsilon_{cvt}
\end{equation}

where $Y_{cvt}$ denotes a review (aggregated) outcome for company $c$, verification status $v$, and date $t$ (which could be either before or after the event). The outcome variables include: (1) the number of reviews, (2) the average rating of the reviews, (3) the number of reviews in each rating category (1–5), (4) the share of reviews in each rating category (1-5), (5) the average number of words per review, (6) a measure of review text homogeneity and (7) three indicator variables for review-volume groups, equal to one when the number of reviews falls into the ranges 0–19, 20–99, and 100 or more, respectively. The variable $After_t$ is a binary indicator equal to one for observations occurring after the LLM supply shocks, within the event window, and zero otherwise. The variable $Treatment_v$ is a binary indicator equal to one for unverified reviews and zero for verified reviews. 

The specification includes company fixed effects ($\alpha_c$) to control for time-invariant differences across companies, and calendar week fixed effects ($\gamma_{w(t)}$) to account for common time shocks affecting all companies. The coefficient $\beta_1$ captures pre-treatment differences between unverified and verified reviews, while $\beta_2$ reflects the average change in outcomes after the LLM supply shock for the reference group (verified reviews). The main parameter of interest, $\beta_3$, measures the differential change over time (after vs. before) in review characteristics for unverified reviews relative to verified reviews, following the OpenAI supply shocks. The error term $\varepsilon_{cvt}$ captures unobserved determinants of the outcome. Standard errors are clustered at the company level, consistent with the assumption that shocks affecting a given company may be persistent, while remaining independent across companies.

\subsection{Event-study specification}\label{subsec:event study}

In addition to the aggregated before--after baseline specification, we estimate an event-study specification that allows the treatment effect to vary across individual days surrounding the LLM supply shocks. Whereas the baseline specification compares review activity over the entire seven-day periods before and after each event, the event-study specification estimates separate effects for each day within the event window.

Specifically, we use daily outcome variables for the fourteen days surrounding each event: seven days before and seven days after the event. This higher-frequency specification allows us to examine the dynamics of review activity within the event window, including whether the response emerges immediately following the event or evolves gradually over subsequent days. It also allows us to assess the identifying parallel-trends assumption by examining differential dynamics between verified and unverified observations during the pre-event period.

Each observation is indexed by company $c$, verification status $v$, and calendar date $t$. Let $\mathrm{Treatment}_v$ be an indicator equal to one for unverified observations and zero for verified observations. We define $D_t^k$ as an indicator equal to one when calendar date $t$ is $k$ days relative to the event date. The event-study window is given by:

\[
k \in \{-7,-6,\ldots,-1,1,\ldots,7\},
\]

where the event day, $k=0$, is excluded.

Rather than selecting a single pre-event day as the reference period, we normalize the coefficients such that their average over the seven pre-event days equals zero. We impose this normalization separately on the event-time coefficients, $\lambda_k$, and the treatment-by-event-time interaction coefficients, $\delta_k$:

\[
\frac{1}{7}\sum_{k=-7}^{-1}\lambda_k=0,
\qquad
\frac{1}{7}\sum_{k=-7}^{-1}\delta_k=0.
\]

Under this normalization, the event-study specification is:

\begin{equation}
\label{eq:eventstudy}
\begin{aligned}
Y_{cvt}
={}&
\alpha_c
+
\gamma_{w(t)}
+
\theta\,\mathrm{Treatment}_v
\\
&+
\sum_{k=-7}^{-2}
\lambda_k
\left(
D_t^k-D_t^{-1}
\right)
+
\sum_{k=1}^{7}
\lambda_k D_t^k
\\
&+
\sum_{k=-7}^{-2}
\delta_k
\mathrm{Treatment}_v
\left(
D_t^k-D_t^{-1}
\right)
+
\sum_{k=1}^{7}
\delta_k
\mathrm{Treatment}_v D_t^k
+
\varepsilon_{cvt}.
\end{aligned}
\end{equation}
Here, $Y_{cvt}$ denotes the outcome for company $c$, verification status $v$, and date $t$; $\alpha_c$ denotes company fixed effects; and $\gamma_{w(t)}$ denotes calendar-week fixed effects, where $w(t)$ is the calendar week containing date $t$. Standard errors are clustered at the company level.

Because the pre-event coefficients are normalized to have mean zero, $\theta$ captures the average difference between unverified and verified observations over the seven pre-event days. The coefficients $\lambda_k$ describe the event-time dynamics for verified observations, the reference group, relative to their average pre-event event-time effect. The interaction coefficients $\delta_k$, which are the primary coefficients of interest, capture how the difference between unverified and verified observations on relative day $k$ deviates from the average difference between the two groups during the seven-day pre-event period.

The coefficients for relative day $-1$, $\lambda_{-1}$ and $\delta_{-1}$, are not estimated directly. Under the normalization above, they are recovered as

\[
\hat{\lambda}_{-1}
=
-\sum_{k=-7}^{-2}\hat{\lambda}_k,
\qquad
\hat{\delta}_{-1}
=
-\sum_{k=-7}^{-2}\hat{\delta}_k.
\]

Consequently, the estimated pre-event coefficients satisfy

\[
\frac{1}{7}\sum_{k=-7}^{-1}\hat{\lambda}_k=0,
\qquad
\frac{1}{7}\sum_{k=-7}^{-1}\hat{\delta}_k=0.
\]

To assess the parallel-trends assumption, we test whether the differential dynamics between unverified and verified observations are jointly zero during the pre-event period. Given the normalization, this is equivalent to testing the six freely estimated pre-event interaction coefficients:

\[
H_0:
\delta_{-7}
=
\delta_{-6}
=
\cdots
=
\delta_{-2}
=
0.
\]

If these six coefficients are jointly equal to zero, the normalization implies that $\delta_{-1}=0$ as well. The joint test therefore evaluates whether there is evidence of systematic differential dynamics between unverified and verified observations prior to the event.

The derivation of the average pre-event normalization and the corresponding reparameterization of the regression specification are provided in Section~\ref{subsec:regression_specification}.

\subsection{Text Homogeneity Measure}\label{subsec:homog}

In addition to standard review-level outcomes, we construct a measure of textual homogeneity that captures the degree of similarity across reviews within a given group. This measure is intended to proxy for the extent to which review content is standardized or exhibits similar linguistic patterns. In other words, it allows us to test whether increased use of generative AI may alter the general textual properties of online reviews. We expect AI-generated reviews to exhibit higher levels of textual homogeneity, following prior research suggesting that AI-generated texts tend to display more uniform linguistic structures, more repetitive lexical patterns, and lower stylistic variation than human-written texts (\cite{kujur2025comparative, culda2025comparative, munoz2024contrasting}).

For each event, we group reviews by company, verification status, and event-time period, distinguishing between reviews posted during the seven days preceding the event and those posted during the seven days following the event. We retain groups containing at least 10 reviews and exclude reviews shorter than 5 characters to remove trivial or non-informative text entries. For groups containing more than 100 reviews, we randomly sample 100 reviews to maintain computational tractability.\footnote{Only 3.34\% of the groups contain more than 100 reviews and therefore require sampling; all remaining eligible groups are used in full.}

We convert each review text into a vector representation using the pre-trained all-MiniLM-L6-v2 Sentence Transformer model. The resulting 384-dimensional embeddings map reviews into a semantic vector space in which reviews with more similar content are located closer to one another. Within each group, we compute a centroid embedding, defined as the average of all review embeddings in that group. For each review, we then calculate its cosine similarity to the group centroid. Higher cosine similarity indicates that a review is more similar to the typical review in its group and, consequently, that the group's review content is more homogeneous.

Finally, for each group, we compute the mean and standard deviation of the cosine similarity scores. The mean captures the overall level of textual homogeneity, whereas the standard deviation captures the dispersion in textual similarity within the group.

\vspace{1\baselineskip}

\section{Results}\label{sec:results}

\subsection{Baseline Difference-in-Differences Analysis}\label{subsec: res baseline}

\subsubsection{The Effect on Rating Distribution}\label{subsubsec: res base rating}

Table \ref{tab:did_week} shows the effect of OpenAI supply shocks on the average rating and on the proportions of 1-star and 5-star ratings. We focus on these three metrics because they provide the clearest evidence of changes in review patterns. We also estimate the same specifications for the 2-, 3-, and 4-star ratings, and also using other standard review metrics, including raw review counts, the logarithm of review counts, review length, and measures of review text homogeneity; however, the estimates of the effects are not statistically significant across specifications. For completeness, the corresponding results are reported in Tables \ref{tab:did_nonsig_explanatory_main} and \ref{tab:did_nonsig_log_counts}.
The coefficient of the \textit{Treatment} variable is large and statistically significant across all specifications, indicating that, prior to the LLM supply shocks, unverified reviews differ from verified reviews in that they exhibit lower average ratings, which is driven by a higher proportion of 1-star ratings and a lower proportion of 5-star ratings. Several mechanisms may explain the greater negativity of unverified reviews, including dissatisfied consumers’ stronger desire for anonymity, positive selection by businesses in encouraging satisfied customers to leave verified reviews, or other unobserved factors. Our empirical setup is designed to account for these baseline differences. The coefficient of the \textit{After} variable is small and not statistically significant across all outcomes, indicating that verified reviews — the reference group — were not significantly affected by the LLM supply shock events. In other words, the behavior of verified reviewers remains broadly stable before and after the shock. 
Turning to the effect of interest, the coefficient of the interaction \textit{Treatment × After} for the average rating is statistically significant. This result indicates that, following OpenAI supply shocks, and given all the controls in this setup, while the average rating did not significantly change for verified reviews, it declined for unverified reviews relative to verified reviews. In other words, we find that a reduction in production costs or an improvement in the capabilities of generative AI models increases the average negativity of ratings directed at businesses on the Trustpilot platform.

\begin{table}[H]
\centering
\caption{Difference-in-Differences Results: Seven-Day Event Window}
\label{tab:did_week}

\begin{threeparttable}

\small
\setlength{\tabcolsep}{4pt}
\renewcommand{\arraystretch}{0.9}

% Columns (1)--(3)
\begin{tabular}{
    p{3.2cm}
    *{3}{>{\centering\arraybackslash}p{3.0cm}}
}
\toprule
 & (1) & (2) & (3) \\
 & Avg. rating & Prop. 1-star & Prop. 5-star \\
\midrule

Treatment & -1.0403*** & 0.2648*** & -0.2268*** \\
 & (0.022) & (0.005) & (0.006) \\

After & 0.0045 & -0.0013 & 0.0006 \\
 & (0.005) & (0.001) & (0.002) \\

Treatment $\times$ After & -0.0171*** & 0.0034*** & -0.0046*** \\
 & (0.005) & (0.001) & (0.001) \\

\midrule
Company FE & Yes & Yes & Yes \\
Week FE & Yes & Yes & Yes \\
Mean (After = 0) & 3.763 & 0.264 & 0.628 \\
R-squared & 0.575 & 0.555 & 0.499 \\
Observations & 748,513 & 748,513 & 748,513 \\
\bottomrule
\end{tabular}

\vspace{0.2cm}

% Columns (4)--(6)
\begin{tabular}{
    p{3.2cm}
    *{3}{>{\centering\arraybackslash}p{3.0cm}}
}
\toprule
  & (4) & (5) & (6) \\
 & Low-volume [0--20) & Moderate-volume [20--100) & High-volume 100+ \\
\midrule

Treatment & 0.001329*** & -0.001155*** & -0.000174*** \\
 & (0.00014) & (0.00013) & (0.00004) \\

After & 0.000234*** & -0.000222*** & -0.000013 \\
 & (0.00003) & (0.00003) & (0.00001) \\

Treatment $\times$ After & -0.000040* & 0.000037 & 0.000002 \\
 & (0.00002) & (0.00002) & (0.00001) \\

\midrule
Company FE & Yes & Yes & Yes \\
Week FE & Yes & Yes & Yes \\
Mean (After = 0) & 0.9987 & 0.0012 & 0.0001 \\
R-squared & 0.268 & 0.244 & 0.216 \\
Observations & 16,116,186 & 16,116,186 & 16,116,186 \\
\bottomrule
\end{tabular}

\vspace{0.1cm}

\begin{tablenotes}[flushleft]
\footnotesize
\item \textit{Notes:} This table reports Difference-in-Differences estimates for review ratings and review-volume categories within the seven-day event window. Columns (1)--(3) report results for the average review rating and the proportions of 1-star and 5-star reviews, respectively. Columns (4)--(6) report results for indicators equal to one if a company--day--verification-status observation belongs to the low-volume (0--19 reviews), moderate-volume (20--99 reviews), or high-volume (100 or more reviews) category, respectively, and zero otherwise. Company--day--verification-status observations with zero reviews are retained in the sample and classified in the low-volume category. The coefficients in Columns (4)--(6) can therefore be interpreted as changes in the probability that an observation belongs to the corresponding review-volume category. The coefficient on the interaction term (\textit{Treatment} $\times$ \textit{After}) captures the differential change following the LLM supply shock for unverified reviews relative to verified reviews. For the rating outcomes, the interaction coefficient is negative and statistically significant for average review rating and the proportion of 5-star reviews, and positive and statistically significant for the proportion of 1-star reviews. For the review-volume categories, the interaction coefficient is negative and marginally significant for the low-volume category, while no statistically significant differential effects are observed for the moderate- and high-volume categories. All regressions include company and week fixed effects. Standard errors are clustered at the company level and reported in parentheses. Statistical significance levels: *** $p<0.01$, ** $p<0.05$, * $p<0.1$.
\end{tablenotes}

\end{threeparttable}
\end{table}

This pattern is consistent with the effect on the distribution of ratings. Specifically, the proportion of one-star reviews increases significantly following LLM supply shocks for unverified reviews relative to verified reviews, with an estimated coefficient of 0.0034 (p-value < 0.01), corresponding to an increase of 0.34 percentage points. Conversely, the proportion of five-star reviews decreases significantly, with a coefficient of -0.0046 (p-value < 0.01), implying a decline of 0.46 percentage points. It is important to note that although the magnitude of these effects may appear small in percentage point terms, they are economically meaningful given the scale of the platform. Even modest changes in proportions translate into a substantial number of reviews when aggregated over millions of observations and for over 72,000 companies and businesses. As such, these shifts can meaningfully influence the overall distribution of ratings, potentially affecting consumer perceptions, firm reputation, and competitive dynamics on the platform.

Two important observations are required to interpret these findings. First, the platform enacts a relatively strict policy of filtering what it believes to be fake reviews (around 6\% are filtered each year). A successful platform-filtering process would presumably leave a negligible amount of AI-generated reviews. Therefore, our results capture the residual AI-generated activity that remains after Trustpilot’s aggressive filtering process has already taken place. Second, from an economic perspective, AI-generated online reviews mostly serve two broad strategic purposes: they can be used either to artificially enhance the reputation of one’s own or affiliated businesses, or to strategically damage competing businesses by generating unfavorable evaluations. Our results suggest that some AI-generated reviews manage to bypass the fake review detection mechanisms of the platform, although we cannot rule out that what we see is a result of some interaction between the platform and the AI-generated reviews activity (e.g., that changes in AI supply affects the filtering process, in some way). Assuming these patterns are indeed AI-generated reviews that evaded platform detection, it seems then that the availability of cheaper or more efficient generative AI tools is used more for competitive strategic purposes rather than self-enhancing. One potential explanation is that artificially inflating one’s own ratings exposes a business to a higher risk of detection and punishment by platform moderation systems, whereas posting negative reviews targeting competitors can achieve a similar effect, by widening the rating gap, while carrying a lower risk of being detected. It may also suggest that AI’s enhanced capabilities make it easier to generate credible negative reviews than positive ones. This strategy may be especially lucrative, given prior evidence that negative reviews exert a stronger influence on economic outcomes. This behavior, if it exists on other platforms, could amplify the competitive dynamics on digital platforms, harm the quality of consumer-generated information on platforms, and increase monitoring costs as firms may increasingly rely on automated content generation to influence perceived product quality and reputation in online review environments.

\subsubsection{The Effect on Review-Volume Groups}\label{subsubsec: res base volume}

As stated in the previous section, the review counts did not exhibit statistically significant effects. Moreover, as shown below in the event-study analysis, the temporal evolution of review counts does not appear to satisfy the parallel-trends assumption. This motivates the question of whether AI-generated reviews are generated in a manner not detectable using a naive counts measure. We therefore investigate an alternative form of review-volume activity: abnormally large numbers of reviews posted during the same day, for a given company and verification status, which may be an indication of coordinated or automated review generation. 

To study this phenomenon, we encode three dummy variables corresponding to different levels of within-day review volume based on the number of reviews received on that date: \textit{low review volume} (0--19 same-company reviews), \textit{moderate review volume} (20--99 same-company reviews), and \textit{high review volume} (100 or more same-company reviews). These thresholds were chosen to capture increasingly high levels of review activity while preserving a sufficient number of observations within each group. We tested several alternative cutoffs, and the results remain qualitatively unchanged.

\begin{table}[H]
\centering

\caption{Descriptive Statistics of Review-Volume Categories}
\label{tab:spike_descriptive_stats}

\begin{minipage}{\textwidth}
\centering

\begin{threeparttable}

\begin{adjustbox}{max width=\textwidth}
\begin{tabular}{lrrr}
\toprule
Statistic
& Low volume [0--20)
& Moderate volume [20--100)
& High volume 100+ \\
\midrule

Number of company--day--verification observations
& 16,095,834
& 18,625
& 1,727 \\

Number of unique companies
& 72,291
& 1,103
& 177 \\

Share of company--day--verification observations (\%)
& 99.87
& 0.12
& 0.01 \\

Total reviews
& 1,716,188
& 685,022
& 395,989 \\

Share of reviews (\%)
& 61.35
& 24.49
& 14.16 \\

Mean reviews per company--day--verification observation
& 0.11
& 36.78
& 229.29 \\

Median reviews per company--day--verification observation
& 0.00
& 31.00
& 159.00 \\

Standard deviation of reviews
& 0.76
& 17.39
& 250.61 \\

25th percentile reviews
& 0.00
& 24.00
& 121.50 \\

75th percentile reviews
& 0.00
& 44.00
& 243.50 \\

Verified reviews (\%)
& 47.38
& 76.57
& 84.87 \\

Unverified reviews (\%)
& 52.62
& 23.43
& 15.13 \\

\bottomrule
\end{tabular}
\end{adjustbox}

\vspace{2pt}

\begin{tablenotes}[flushleft]
\footnotesize
\item[]
\parbox{\textwidth}{%
\textit{Notes:}
This table reports descriptive statistics for the three review-volume
categories used in the analysis. The unit of observation is a
company--day--verification-status observation within the
event-window panel. Low-volume observations correspond to 0--19 reviews,
moderate-volume observations correspond to 20--99 reviews, and high-volume
observations correspond to 100 or more reviews. Company--day--verification-status
observations with zero reviews are retained in the sample and classified in
the low-volume category. Verified and unverified review percentages are
calculated based on the number of reviews within each review-volume category.
The sample is restricted to review dates in 2023--2024.
}
\end{tablenotes}

\end{threeparttable}

\end{minipage}
\end{table}

Table~\ref{tab:spike_descriptive_stats} reports descriptive statistics for the three review-volume categories. The descriptive statistics are based on the panel used in the analysis and therefore include company--day--verification-status observations with zero reviews, which are classified in the low-volume category. Consequently, low-volume observations account for 99.87\% of all company--day--verification-status observations but only 61.35\% of total reviews, reflecting that most observations contain few or no reviews. In contrast, moderate- and high-volume observations are relatively rare, representing only 0.12\% and 0.01\% of all observations, respectively, yet together account for nearly 39\% of all reviews in the sample. High-volume observations are particularly concentrated: although they comprise only 0.01\% of company--day--verification-status observations, they account for 14.16\% of all reviews. The table also shows that verified reviews are disproportionately represented among moderate- and high-volume observations, whereas unverified reviews are relatively more prevalent among low-volume observations.

The estimation results for these three review-volume outcome variables are reported in the second panel of Table \ref{tab:did_week}. \footnote{We also estimated nonlinear specifications (logit and binomial models) for these outcomes. However, due to the inclusion of a large number of fixed effects (e.g., company and time effects), these models face substantial econometric and computational challenges. In particular, fixed effects estimators in nonlinear panel models such as logit and probit are known to suffer from the incidental parameters problem, which can lead to severe bias and unreliable inference (\cite{cruz_gonzalez_2017, hahn_kuersteiner_2011}). As a result, we adopt a linear fixed effects specification, which remains computationally tractable and provides a reasonable approximation to average partial effects (as discussed in \cite[p.~563]{wooldridge_2010}).}

As in the previous specifications, the coefficient of the \textit{Treatment} indicator shows that the distribution of review volume differs systematically between unverified and verified reviews in the pre-treatment period. In particular, low-volume days (0--19 reviews) are more likely to appear for unverified reviews, while moderate- and high-volume days are less likely. This suggests a lower prevalence of high-volume review activity among unverified reviews, which could reflect systematic differences between the two types of reviews or, alternatively, stronger platform filtering of unverified reviews.

The estimated coefficients of the interaction term \textit{Treatment × After} for the moderate- and high-volume categories are not statistically significant. On the other hand, we do find evidence of a marginally significant \textit{negative} effect for the low-volume category (0--19 reviews) at the 10\% significance level. This effect can be interpreted as the equivalent of an \textit{increase} in the complementary review-volume category, namely days with 20 or more same-day reviews (i.e., moderate- and high-volume days), following LLM supply shocks. Since batches of more than 20 unverified reviews in a single day for a single company appear to be mostly unlikely under normal conditions, we interpret this as evidence of AI-generated review activity that succeeds in bypassing platform filters. In the next section, we present additional evidence at the daily level suggesting that AI-generated activity may manifest itself through unusually high review-volume days.

\enlargethispage{\baselineskip}
\subsection{Event Study Specification}\label{subsec: res event study}

As described in Section \ref{sec:methods}, we further estimate a difference-in-differences specification with company and calendar-week fixed effects, but instead of using a single indicator for the entire post-event window, the regression includes separate indicators for each day before and after the LLM supply shock date. Rather than using a single pre-event day as the omitted reference category, the specification normalizes all event-time coefficients relative to the average of the seven pre-treatment days. Accordingly, the estimated coefficients capture deviations from the average pre-treatment period. The analysis considers the same six outcome variables examined in the previous subsection.

\begin{figure}[H]
    \centering

    \caption{Event-Study Estimates of the Effect of LLM Supply Shocks on Rating Outcomes and Review-Volume Categories}
    \label{fig:event_study_combined}

    \begin{minipage}{\textwidth}
        \centering
        \includegraphics[width=\textwidth]{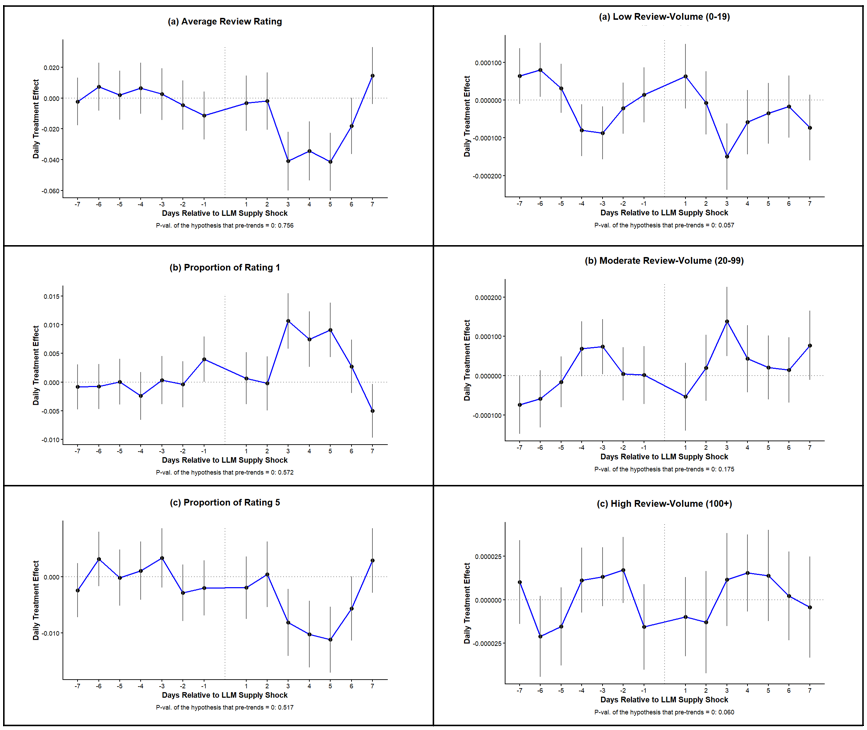}

        \vspace{0.3em}

        {\scriptsize
\justifying
        \noindent\textit{Notes:} This figure reports event-study estimates of the effect of LLM supply shocks on rating outcomes and review-volume categories. The left column reports rating outcomes: Panel (a) shows the average review rating, Panel (b) the proportion of one-star reviews, and Panel (c) the proportion of five-star reviews. The dependent variables in the right column are indicators equal to one if a company--day--verification-status observation belongs to the low-volume (0--19 reviews) category in Panel (a), the moderate-volume (20--99 reviews) category in Panel (b), or the high-volume (100 or more reviews) category in Panel (c), and zero otherwise. The coefficients can therefore be interpreted as changes in the probability that an observation belongs to the corresponding review-volume category. The sample consists of aggregated observations at the company--day--verification-status level within a symmetric 7-day window before and after each LLM supply shock (see Section~\ref{sec:data_preparation} for details on data construction). The treatment group comprises unverified reviews, while verified reviews serve as the control group. The coefficients are obtained from the event-study specification in Equation~(\ref{eq:eventstudy}), where each outcome is regressed on interactions between an indicator for unverified reviews and a set of event-time dummies. The event-time coefficients are normalized such that their average over the seven pre-treatment days ($-7$ through $-1$) equals zero. To implement this normalization, the coefficient for day $-1$ is omitted from the regression and subsequently reconstructed as the negative sum of the estimated coefficients for the other six pre-treatment days. Each coefficient represents the estimated differential change in the outcome for unverified relative to verified reviews at that event time, expressed relative to the average differential during the pre-treatment period. All regressions include company fixed effects and calendar week fixed effects. Standard errors are clustered at the company level, and error bars represent 95\% confidence intervals. The vertical dashed line indicates the timing of the LLM supply shock (day 0). The p-value reported below each panel corresponds to a joint test of the null hypothesis that the six freely estimated pre-treatment coefficients are jointly equal to zero. The rating-outcome panels are based on $N = 748{,}513$ observations. The review-volume panels are based on $N = 16{,}116{,}186$ observations. This number exceeds that for the rating outcomes because company--day--verification-status observations with zero reviews are coded as zeros in the review-volume category indicators and therefore included in the review-volume analyses, whereas they are treated as missing in the rating and proportion analyses.
        \par}
    \end{minipage}

\end{figure}

Figure \ref{fig:event_study_combined} presents the estimated interaction-term coefficients of the event-study specification for the six outcomes. The left column includes figures for which the dependent variable is the following rating outcomes: average review rating, the proportion of one-star reviews, and the proportion of five-star reviews. The right column includes the figures for which the dependent variable is the surge review-volume categories: low, moderate, and high review-volume.

The review rating outcome panels in the left column reveal several noteworthy patterns. The estimated daily coefficients in the pre-treatment period are generally small and statistically indistinguishable from zero. Moreover, the formal pre-trend tests fail to reject the null hypothesis of no differential pre-treatment trends for all three outcome variables, providing support for the parallel trends assumption (average rating: pre p-value = 0.756; one-star proportion: pre p-value = 0.572; five-star proportion: pre p-value = 0.517). We do see a statistically significant pattern for the post-event dynamics. In the days following the LLM supply shocks, the average review rating initially declines before gradually recovering, with the effect being most pronounced approximately three to five days after the LLM supply shocks. This pattern appears to be driven by an increase in the relative share of one-star reviews during the seven days following LLM supply shocks, accompanied by a corresponding decrease in the share of five-star reviews. We do not detect meaningful changes in the other rating categories (2, 3, and 4), suggesting that any AI-generated activity may occur primarily within the one-star or five-star categories. These dynamic responses are consistent with the aggregated seven-day results and, taken together, suggest that after price reductions and new model launches, there seems to be a shift of AI-generated reviews usage from positive self-promotion to negative competitive efforts. At the same time, the results reveal the localized temporal structure of the suspected AI-generated activity, indicating that the activity is done in a "concentrated wave" rather than diffusely. We do note that this pattern could be either due to producers’ strategic timing or to some sort of interaction between review generation and the platform’s filtering mechanisms. Given the nature of the research setup and the data, we are unable to discern between the two scenarios. Notably, these effects are averaged across a highly diverse set of categories and over 72,000 companies and businesses.

The review-volume panels in the right column present the corresponding event-study estimates for the three review-volume surge categories at the daily level, i.e., a cluster of same-company-same-day reviews. Specifically, we focus on three categories of per-day review volume: low-volume days (0--19 reviews per day), moderate-volume days (20--99 reviews per day), and high-volume days (100 or more reviews per day), for the same company.

First, we find that the pre-treatment coefficients satisfy the parallel trends condition for the moderate-volume category (pre p-value = 0.175). For this category, we also observe statistically significant post-treatment dynamics (post p-value = 0.022). Similarly to the review rating and distribution, here too the effect peaks approximately three days after the event in the form of a temporal "wave," and is broadly consistent with the aggregate-analysis results.

For the other two categories, the pre-trend tests are not statistically significant at the conventional 5\% level, but remain relatively close to the threshold and therefore make it difficult to fully rule out some degree of pre-treatment dynamics (low-volume days: pre p-value = 0.057; high-volume days: pre p-value = 0.060). For low-volume days, we detect statistically significant post-treatment dynamics (post p-value = 0.009), while for high-volume days the post-treatment effects are not statistically significant (post p-value = 0.570).

Overall, the findings regarding the same-day-same-company review surges suggest that at least part of the AI-generated review activity that escapes platform filtering may occur in the form of moderate-volume review days and, to a lesser extent, low-volume review days. More broadly, the results indicate that suspected AI-generated reviews in some cases tend to emerge in temporally concentrated bursts.

\subsection{Review-Volume Effects Among Extreme Ratings}
\label{subsec:which_volume}

To better understand which reviews drive the observed changes in review volume, we focus on the most extreme ratings: 1-star and 5-star reviews. These categories account for the majority of reviews in our sample and are particularly relevant because our previous analyses show that the most pronounced rating effects are concentrated at the extremes of the rating distribution. We therefore examine the combined volume of 1-star and 5-star reviews.

Because review-count distributions differ across rating groups, we adjust the volume thresholds using the same quantile cutoffs as in the baseline specification. For the combined 1-star and 5-star distribution, this yields three categories: low volume (0--17 reviews), moderate volume (18--84 reviews), and high volume (85 or more reviews).

Table~\ref{tab:did_r1r5_adjusted} reports the baseline difference-in-differences estimates for these categories. The \textit{Treatment} $\times$ \textit{After} coefficients are not statistically significant in any of the three specifications: $-0.000035$ for low volume, $0.000036$ for moderate volume, and $-0.000001$ for high volume. Thus, the baseline specification provides no evidence of a statistically significant differential post-shock change between unverified and verified reviews in any of the three combined 1-star and 5-star volume categories.

The event-study results in Figure~\ref{fig:event_study_r1_r5_adjusted}, however, reveal short-run dynamics that are not captured by the baseline specification. The joint post-treatment test rejects the null of no post-treatment effects for the low-volume category (p-value = 0.021) and provides weaker evidence for the moderate-volume category (p-value = 0.096), while the high-volume category is not significant (p-value = 0.178). The parallel-trends assumption is supported for the low- and moderate-volume categories, with pre-trend p-values of $0.134$ and $0.600$, respectively. By contrast, the high-volume category exhibits a significant pre-trend (p-value = 0.013), making its post-treatment dynamics less easily interpretable.

The difference between the baseline and event-study findings reflects the temporal aggregation imposed by the baseline specification. The baseline model estimates a single \textit{Treatment} $\times$ \textit{After} coefficient across the seven-day post-treatment period, whereas the event study allows the treatment effect to vary by day. Short-lived effects can therefore be obscured when aggregated across the full post-treatment window, particularly when effects differ in magnitude or direction across days.

This pattern is visible in Figure~\ref{fig:event_study_r1_r5_adjusted}. For the low-volume category, the treatment effect is close to zero during the first two post-treatment days, falls sharply on day 3, and then gradually returns toward zero. The moderate-volume category exhibits the opposite pattern: the effect rises sharply on day 3 and subsequently declines toward zero. Thus, the third post-treatment day is characterized by a simultaneous decline in low-volume extreme-rating activity and an increase in moderate-volume extreme-rating activity.

These dynamics are consistent with the main event-study results in Figure~\ref{fig:event_study_combined}. While the main analysis documents the broader response in review activity following LLM supply shocks, the results here suggest that an important part of this response is concentrated among 1-star and 5-star reviews. The category-specific analysis therefore suggests that the overall review-volume response is partly driven by a short-run increase in the concentration of 1-star and 5-star reviews.

One possible interpretation is that access to lower-cost and more capable LLMs facilitates short-lived bursts of strategically valuable review generation. The movement from low- to moderate-volume 1-star and 5-star activity around day 3 is consistent with a temporary increase in the concentration of extreme reviews. Five-star reviews could potentially be used to improve a firm's own reputation, whereas 1-star reviews could be directed toward competitors. Because these ratings lie at the extremes of the distribution, they may be particularly effective at influencing ratings and consumer perceptions. This interpretation is suggestive rather than causal evidence of firms' intentions, as the data identify changes in review activity but not who generated individual reviews or for what purpose.

\subsection{Heterogeneity Across AI-Related Events - Which Type is Driving the Effects?}\label{subsec:heterog}

We are interested in identifying which types of events drive the effects we observe: price reductions, new model releases, or events that combine both. To examine this type of heterogeneity in the effects of AI-related shocks, we estimate both the baseline Difference-in-Differences specification and the event-study specification separately for three categories of LLM supply shocks: (i) price reduction events only, (ii) new model release events only, and (iii) events that involve both price reductions and new model releases. Table~\ref{tab:openai_price_reductions} presents the timeline of OpenAI pricing reductions and new model releases used to construct these event categories, including the corresponding event dates and shock types.

We begin by examining the baseline Difference-in-Differences estimates. The results are reported in Tables~\ref{tab:price_reduction_dates}, \ref{tab:new_model_release_dates}, and \ref{tab:price_reduction_and_new_model_release_dates}. The estimates reveal substantial heterogeneity across AI-related event types. For price reduction events and events that combine price reductions with new model releases, the interaction coefficients are generally small and statistically insignificant across all outcomes. In contrast, the interaction coefficients associated with new model release events are statistically significant for most outcomes and closely resemble the patterns observed in the main specification that pools all event types together. Specifically, new model releases are associated with a significant decline in the average review rating and in the proportion of 5-star reviews, alongside a significant increase in the proportion of 1-star reviews. At the same time, the probability of observing low-volume review days (0--19 reviews) declines significantly, while the probability of observing moderate-volume review days (20--99 reviews) increases significantly. These findings suggest that the main effects documented in the baseline Difference-in-Differences analysis are largely concentrated around new model release dates rather than price-reduction events. To investigate the dynamics underlying these effects, we next estimate event-study specifications separately for each event category.

\begin{figure}[H]
\centering
\caption{Event-Study Estimates for Price Reduction Events Only}
\label{fig:price_reduction_events}
\includegraphics[width=\textwidth]{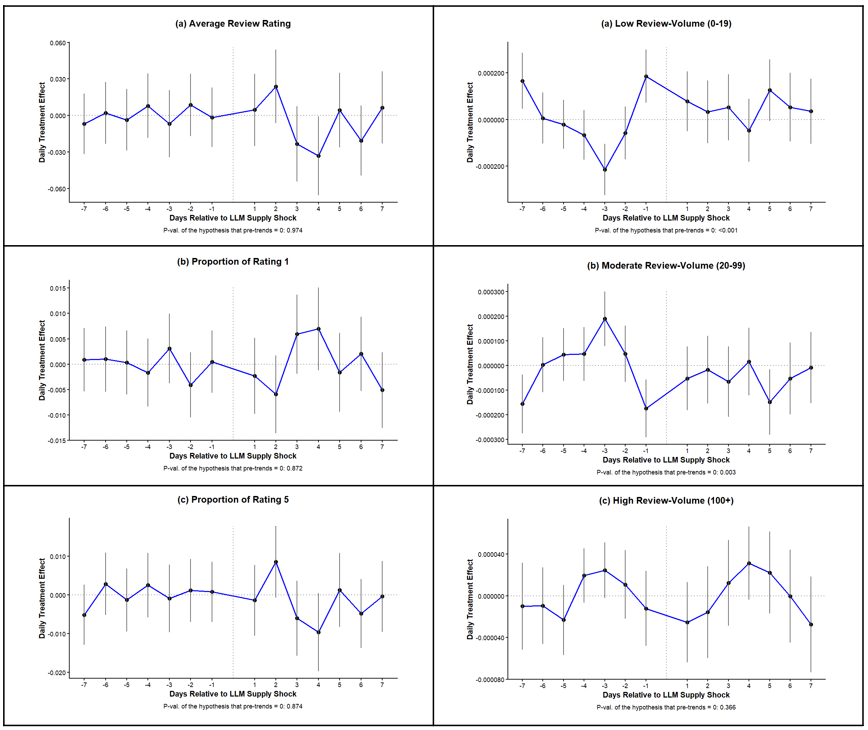}
{\scriptsize
\justifying
\noindent\textit{Notes:} This figure follows the same design and specification as Figure~\ref{fig:event_study_combined}, but restricts the analysis to LLM supply shocks consisting exclusively of price reductions (see Table~\ref{tab:openai_price_reductions}). The rating-outcome panels are based on $N = 286,490$ observations, and the review-volume panels are based on $N = 6,010,256$ observations.
\par}
\end{figure}

\begin{figure}[H]
\centering
\caption{Event-Study Estimates for New Model Release Events Only}
\label{fig:new_model_events}
\includegraphics[width=\textwidth]{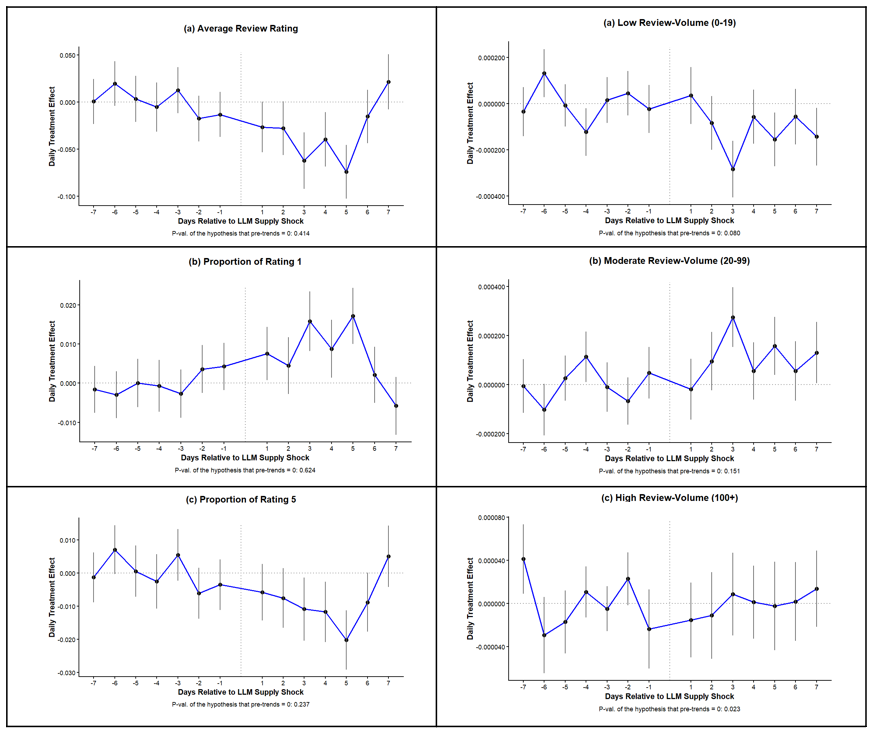}
{\scriptsize
\justifying
\noindent\textit{Notes:} This figure follows the same design and specification as Figure~\ref{fig:event_study_combined}, but restricts the analysis to LLM supply shocks consisting exclusively of new model releases (see Table~\ref{tab:openai_price_reductions}). The rating-outcome panels are based on $N = 308,237$ observations, and the review-volume panels are based on $N = 7,200,354$ observations.
\par}
\end{figure}

\begin{figure}[H]
\centering
\caption{Event-Study Estimates for Events Involving Both Price Reductions and New Model Releases}
\label{fig:both_events}
\includegraphics[width=\textwidth]{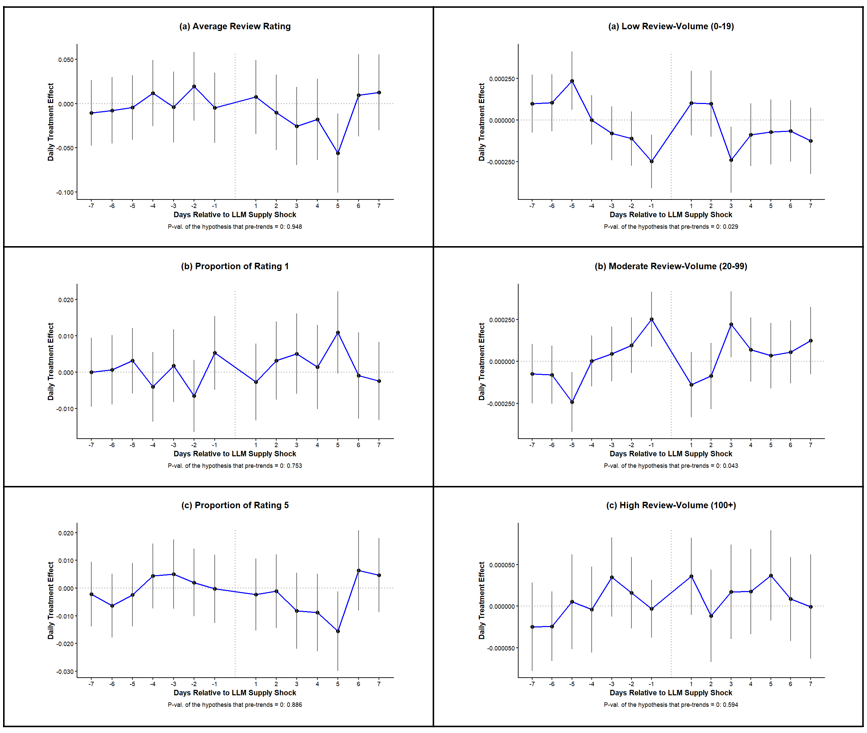}
{\scriptsize
\justifying
\noindent\textit{Notes:} This figure follows the same design and specification as Figure~\ref{fig:event_study_combined}, but restricts the analysis to LLM supply shocks involving both a price reduction and a new model release (see Table~\ref{tab:openai_price_reductions}). The rating-outcome panels are based on $N = 130,283$ observations, and the review-volume panels are based on $N = 2,905,576$ observations.
\par}
\end{figure}

The results of the event-study analysis separated into the three categories are shown in Figures~\ref{fig:price_reduction_events}, \ref{fig:new_model_events}, and \ref{fig:both_events}.

The results are consistent and further reveal substantial heterogeneity across the types of AI shocks. For price reduction events, the evidence is generally weak and does not support robust post-event effects. Specifically, in terms of rating-related outcomes, namely \textit{Average Review Rating}, \textit{Proportion of Rating 1}, and \textit{Proportion of Rating 5}, the post-period p-values are all statistically insignificant, indicating no meaningful post-event changes in the firm rating distributions (see Table~\ref{tab:event_study_ratings}). Notably, the pre-period p-values for these outcomes show statistical insignificance, suggesting that the parallel trends assumption is satisfied but that the treatment effects themselves are weak or absent. In terms of review-volume surge variables, the results are similarly weak, and we do not observe notable effects for any of the review-volume outcomes. Figure~\ref{fig:both_events} shows a similar pattern for events that combine price reductions and new model releases: we do not observe notable effects for any of the outcomes examined.

In contrast, we observe robust effects for the new model release events. For all three rating-related outcomes, the parallel trends assumption holds and the post-period effects are significant. These findings indicate that new model releases are likely driving the effects observed in firms' rating distributions. They suggest that producers of AI-generated reviews respond primarily to changes in model capabilities and efficiency rather than to reductions in the monetary costs of using the models. Interestingly, we do not see effects for the two events which include both price reductions and new model releases. Although the very small number of events prohibits any credible conclusion, we hypothesize that the null-result models (GPT-4-Turbo-128K and text-embedding-3-small, see Table \ref{tab:openai_price_reductions}) differ from the others in ways that potentially make them less likely to affect review production: text-embedding-3-small is not a generative model and therefore cannot produce review text, while GPT-4 Turbo was a relatively costly, resource-intensive model compared with the newer, faster, and cheaper GPT-4o-based models. Therefore, we loosely hypothesize that the absence of detectable effects for these releases may be consistent with limited relevance to large-scale review generation, whereas the remaining models more directly reduced the cost or increased the accessibility and efficiency of producing such content.

The dynamic pattern is similar to what we observed across all events in Figure~\ref{fig:event_study_combined}. Both the \textit{Average Review Rating} and the \textit{Proportion of Rating 1} initially decline shortly after the LLM supply shock and then subsequently increase. In contrast, the \textit{Proportion of Rating 5} exhibits the opposite pattern, increasing immediately following the LLM supply shock before gradually declining.

The review-volume outcomes are also consistent with this interpretation. Both the low-volume and moderate-volume outcomes satisfy the parallel trends assumption and exhibit highly significant post-period effects (p-value < 0.001 for both variables; see Table~\ref{tab:event_study_reviewcounts}). Interestingly, the dynamic patterns displayed in Figure~\ref{fig:new_model_events} differ across these two outcomes. For the low-volume category, the estimated effect declines immediately after the new model release and then gradually increases. In contrast, the moderate-volume category exhibits a sharp increase immediately after the release, followed by a gradual decline.

Taken together, these results suggest that improvements in GenAI capabilities and efficiency associated with the release of new models may increase firms' incentives or ability to generate AI-assisted reviews, whereas reductions in model prices alone appear to have little impact on review-generation behavior.

\section{Robustness and Further Heterogeneity Checks}\label{sec:robust}

\subsection{Are the Effects Driven by a Specific Event?}\label{subsec:separate dates}

To further assess the robustness of our findings and examine whether they are driven by a unique LLM supply shock on a specific date, we re-estimate the event-study specification separately for each event date. Tables~\ref{tab:event_study_ratings} and~\ref{tab:event_study_reviewcounts} present the event-level diagnostics for the rating-related and review-count outcomes, respectively. 

Several patterns emerge from the event-level analysis. First, the effects are not concentrated on a single event but appear across several event dates. In particular, significant post-period effects combined with non-significant pre-trends are observed for the price reduction on 2023-03-01 and across several subsequent events, including multiple new model releases. For the 2023-03-01 price reduction, significant post-event dynamics are observed for average ratings and the share of 5-star ratings, with no evidence of differential pre-trends for these outcomes. Significant post-period effects with non-significant pre-trends also arise for several outcomes around the new model release events on 2024-05-13, 2024-07-18, and 2024-09-12.

Interestingly, the effects within the review-volume categories appear weaker when the analysis is conducted separately by LLM supply shock date than in the pooled specification. A likely explanation is that moderate- and high-volume review days are relatively rare by construction, reducing the number of observations available for each individual event and consequently lowering statistical power. Evidence of higher-volume review activity is present for some event dates, but the effects are generally sparse and not consistently observed across LLM supply shocks.

Overall, event-level analysis supports our interpretation of the main findings by showing that the phenomenon occurs across multiple points in time and across multiple events, suggesting that they reflect a broader pattern related to the usage of AI on the platform rather than stemming from a singular event.

\subsection{Heterogeneity of the Effect by Company Market Size}\label{subsec: robust company size}

Because firms of different sizes may differ in both their incentives and capacity to adopt AI for review generation, we next examine heterogeneity by firm size to determine whether the observed effects are concentrated among smaller or larger firms. Specifically, we examine whether the estimated treatment effects vary by company size, proxied by the volume of company-related activity on the platform. For simplicity, we assume that the total number of reviews (both verified and non-verified) received during 2023--2024 on the platform can be used as a proxy for the relative rank of the company's market size on and outside the platform. 

It is important to note that given that our model uses company fixed effects, small firms with only a handful of reviews do not contribute to the diff-in-diff analysis, due to insufficient observations and lack of variance. This affects how we partition firm market size categories. We therefore restrict the sample to companies that received at least 50 reviews during the sample period. These companies are then divided into four equally sized quartiles based on their total review volume and we estimate our baseline model (Equation ~\ref{eq:did}) separately for each quartile. Table \ref{tab:size_quantiles_summary} reports the review-volume ranges associated with each quartile, while Tables \ref{tab:q1} -- \ref{tab:q4} present the corresponding regression results.

\begin{table}[H]
\centering
\caption{Company Size Quantiles Based on Total Review Volume}
\label{tab:size_quantiles_summary}
\begin{tabular}{lcc}
\toprule
Quantile & Review-volume range & Number of companies \\
\midrule
Q1 (Lowest)  & 50--80        & 5,056 \\
Q2           & 80--152       & 5,056 \\
Q3           & 152--431      & 5,056 \\
Q4 (Highest) & 431--240,745  & 5,056 \\
\bottomrule
\end{tabular}
\end{table}

The results reported in Tables \ref{tab:q1} -- \ref{tab:q4} suggest that the relationship between the treatment and review outcomes varies across company-market-size groups. The largest effects are observed among companies within both the highest and lowest review count quartiles. In both quartiles, the coefficient of the \textit{Treatment} $\times$ \textit{After} interaction is negative and statistically significant for both the average review rating and the share of 5-star reviews. Even though the effect on 1-star reviews is weak in this case, given that we look at the share of review stars the effect we are observing in this case is consistent with the general effect we see in other specifications of movement between positive and negative reviews. 
This suggests that AI-related activity is concentrated among very small firms and firms with exceptionally high levels of activity, but not among firms with moderate levels of activity. Although identifying the precise mechanism is beyond the scope of this study, one possible explanation is that very small and very large firms may represent more attractive targets in terms of bang-for-the-buck adversarial AI usage. The potential harm caused by a negative review depends both on the number of existing reviews and on the firm’s market size. In the perspective of the aggressor AI-using firms: if their competitor is a very small firm or business, even a small number of AI-generated negative reviews can meaningfully damage reputation on the platform, and in general. If the competitor is a large, high-demand firm, pushing even a small fraction of customers to switch firms may be of value to the aggressor firm. 

Finally, in terms of review-volume surge activity, as in the previous heterogeneity analyses, the relative rarity of moderate- and high-volume review days limits statistical power, which may explain why in this case we find no meaningful effects.

\section{Discussion, Limitations and Future Research}
\label{sec:conclusion}
At least in the context examined here, this study provides causal evidence that advances in generative artificial intelligence are beginning to reshape online information ecosystems by changing the economics of textual content production. Using OpenAI's API pricing reductions and new model releases as exogenous shocks, we find systematic changes in unverified reviews on Trustpilot following improvements in AI capabilities and accessibility.

First and foremost, a central contribution of this study is to provide evidence that AI is being used systematically within a marketplace, likely in pursuit of economic objectives. It is important to note that because our approach is designed to detect changes in review activity in response to LLM supply shocks, it captures only a lower bound of the underlying AI-related activity. Furthermore, our findings suggest that the rapid advances of large language models, effectively providing "intelligence as a service" for content creation, are reshaping firms’ incentives to strategically manipulate online information, with potential consequences for the economics of digital platforms. Among other implications, these findings suggest that the availability of AI tools may be altering the dynamics of competition in the relevant markets.

Our findings contribute to the growing literature on AI-generated reviews and AI-generated content in a notable way. Most existing studies have focused on distinguishing AI-generated reviews or content from authentic human reviews by identifying linguistic characteristics or developing automated detection methods (\cite{zhao2025ai, fariello2024distinguishing, guo2024detective, chaka2024reviewing}). Although these studies have improved our understanding of how AI-generated reviews or content differ from human-written content, they provide limited evidence on whether firms actually change their behavior as generative AI becomes more accessible, and in what manners. By leveraging exogenous reductions in the cost and improvements in the capabilities of state-of-the-art language models, our study instead examines the behavioral consequences of generative AI adoption in a real-world marketplace. This perspective complements existing research by shifting attention from identifying AI-generated reviews and content to understanding how advances in generative AI influence firms' strategic behavior.

A robust finding across our analyses is that the observed effects are mainly associated with negative reviews activity. Analogous to other marketing-oriented uses of AI, firms might reasonably be expected to use generative AI to enhance their own reputations by producing favorable reviews that appear authentic, particularly because doing so is relatively inexpensive and requires little effort compared with, e.g., relying on human labor or experts. Instead, our findings are more consistent with generative AI being used as a competitive tool that enables firms to target rivals more easily. This emerging development may have important implications for the competitive structure of markets. Although our empirical design does not allow us to identify the motivations of reviewers, several mechanisms may explain this pattern. Positive AI-generated self-promotion may expose the perpetrator to greater risk if detected by the platform, because the identity of the benefiting firm is readily apparent. This risk is further amplified if platforms can detect AI-generated self-promotion more effectively than negative attacks on competitors. Under such conditions, improvements in AI’s ability to produce credible, authentic-looking negative content may make such manipulation a relatively less risky and potentially more effective strategy. This possibility becomes even more salient given that damaging a rival’s reputation may yield greater returns per AI-generated review, as prior research suggests that negative reviews exert a stronger influence than positive ones (\cite{chevalier2006effect}). Finally, fabricating a negative review that contains detailed, well-articulated criticisms of a business may be more difficult than generating a generic positive review, making such reviews more dependent on the enhanced capabilities of advanced AI models. Taken together, these considerations suggest that generative AI may not only increase the volume of manipulated content but also shift the strategic direction of manipulation itself.

Our results further suggest that advances in intelligence as a service, the release of new, more capable models, matter more than API price reductions alone. Although our empirical strategy is motivated partly by leveraging declines in the effective cost of AI-generated text, the strongest effects coincide with model releases, indicating that firms may be more responsive to improvements in quality, reasoning, contextual understanding, and production efficiency than to lower monetary costs. In simpler terms, this indicates that firms value more capable AI models. Consequently, as a small number of providers increasingly supply “intelligence as a service,” AI capabilities may become more concentrated in the hands of a few firms, raising concerns about market power and dependence on these providers.

Another contribution concerns the temporal organization of suspected AI-generated review activity. Our approach focuses on short-term responses and does not capture persistent longer-run effects. The review-volume and event-study results suggest that at least some AI-related activity occurs in concentrated bursts around AI supply shocks rather than as a continuous increase in review production. Also, the rise in review surges following LLM supply shocks is consistent with coordinated waves of activity and indicates that users of AI-generated reviews adapt quickly to changes in model availability. This temporal concentration may also provide a useful signal for detecting similar activity in future work.

Our heterogeneity analyses further suggest that these strategic incentives vary across firm activity levels. The estimated effects are strongest among firms with very small and very large review volumes, consistent with the possibility that the expected returns to manipulation depend on market position. Although this interpretation remains speculative, both groups may represent especially attractive targets for AI-generated attacks: a small number of strategically generated reviews can materially affect firms with limited review histories, while highly visible firms may be targeted because even modest reputational changes can influence a large customer base. This finding is consistent with the general interpretation of our findings that generative AI is being deployed as a strategic tool.

A major alternative explanation is that LLM supply shocks induce changes in the platform’s filtering process rather than in AI-generated review activity. While platform responses are likely slower than those of firms or contractors, the findings themselves also make a simple platform-wide filtering explanation less plausible. Effects concentrated among the lowest- and highest-activity firms and short-lived same-company review bursts are more consistent with targeted AI-generated review activity than with a uniform moderation change. Although we cannot rule out filtering changes that interact with firm characteristics or AI capabilities, such an explanation would require a more specific moderation response.

This study has several limitations. First, our analysis focuses on a single online review platform. Although Trustpilot is one of the largest review platforms which covers a very wide range of categories and companies, and provides institutional features that are well suited to our identification strategy, future research should examine whether similar patterns emerge on other platforms with different moderation policies, verification mechanisms, and user populations. Second, our empirical strategy provides indirect evidence regarding AI-assisted review generation changes rather than direct identification of the \textit{total} amount of AI-generated reviews. Third, our approach is better suited to identifying short-term effects of LLM supply shocks on online review dynamics than longer-term effects, limiting our ability to assess their persistence and broader long-run consequences, which we recommend future research examine. Fourth, while exploiting exogenous variation generated by LLM supply shocks offers important advantages for causal inference, our design does not allow us to identify the actors responsible for generating the new reviews. For example, whether they are produced by the interested firms themselves or by hired specialized contractors with the ability to produce AI-generated content, or whether the findings we observe are a result of some sort of interaction with the platform's filtering mechanisms. Fifth, our analysis relies on the assumption that verified reviews constitute an appropriate control group because they are substantially less susceptible to strategic manipulation than unverified reviews. Although we are relatively confident that verified reviews are very hard to manipulate at scale using AI models, as we have seen in the estimations, in some cases they may not be a perfect control for the unverified reviews. Interestingly, they appear to perform better as control variables for normalized measures of review activity, such as rating shares, possibly because normalization helps account for underlying fluctuations and differences between the overall volumes of verified and unverified reviews.

An additional limitation is that the Trustpilot data include only reviews that remain after the platform’s moderation and filtering procedures have been applied. According to Trustpilot's transparency reports, reviews identified as fake are removed before becoming part of the data analyzed in this study. Consequently, our estimates capture only the AI-related review activity that remains observable after platform moderation. This implies that the observed AI-related effects represent a lower bound of the actual AI-related activity and that the behavioral changes we document persist despite the presence of sophisticated review-filtering mechanisms.

Finally, our findings suggest several directions for future research. Applying similar identification strategies across review platforms, social media, and digital marketplaces could provide an approach that does not rely on text-based markers or classification methods, helping clarify the generalizability of these patterns and the extent to which they depend on institutional settings and moderation policies.
Examining the developments of other AI providers, such as Google, Anthropic, and Meta, would help determine whether the observed responses reflect broader changes in the generative AI ecosystem, and how they differ across AI tools. Similar mechanisms may extend beyond online reviews to broader forms of user-generated content that have become critical components of digital platforms and the modern economy. These include social media posts, online forums, question-and-answer communities, and recommendation systems, where AI-generated content may similarly influence information flows and competitive dynamics. Finally, future work should examine how platforms adapt to increasingly capable generative AI and how advances in content generation, strategic firm behavior, and moderation technologies jointly shape the credibility of digital information ecosystems.

\clearpage
\bibliographystyle{apalike}
\bibliography{references}

@online{trustpilot_transparency,
  author  = {{Trustpilot}},
  title   = {Transparency Report},
  year    = {2024},
  url     = {https://corporate.trustpilot.com/press/news/transparency-report},
  note    = {\url{https://corporate.trustpilot.com/press/news/transparency-report} Accessed: 2026-03-12}
}

@online{openai_forum_cost_comparison,
  author  = {{OpenAI Developer Community}},
  title   = {GPT-4 and GPT-3.5 Turbo API Cost Comparison and Understanding},
  year    = {2023},
  url     = {https://community.openai.com/t/gpt4-and-gpt-3-5-turb-api-cost-comparison-and-understanding/106192},
  note    = {Accessed: 2026-03-12}
}

@online{voicebot_openai_price_reduction_2023,
  author  = {Schwartz, Eric Hal},
  title   = {OpenAI Upgrades GPT-4 and GPT-3.5 Turbo Models, Reduces API Prices},
  year    = {2023},
  url     = {https://voicebot.ai/2023/06/13/openai-upgrades-gpt-4-and-gpt-3-5-turbo-models-reduces-api-prices/},
  note    = {Voicebot.ai. Accessed: 2026-03-12}
}

@online{openai_devday_2023,
  author  = {{OpenAI}},
  title   = {New Models and Developer Products Announced at DevDay},
  year    = {2023},
  url     = {https://openai.com/index/new-models-and-developer-products-announced-at-devday/},
  note    = {Accessed: 2026-03-12}
}

@online{openai_embedding_api_updates_2024,
  author  = {{OpenAI}},
  title   = {New Embedding Models and API Updates},
  year    = {2024},
  url     = {https://openai.com/index/new-embedding-models-and-api-updates/},
  note    = {Accessed: 2026-03-12}
}

@online{openai_hello_gpt4o_2024,
  author  = {{OpenAI}},
  title   = {Hello GPT-4o},
  year    = {2024},
  url     = {https://openai.com/index/hello-gpt-4o/},
  note    = {Accessed: 2026-03-12}
}

@online{openai_gpt4o_mini_2024,
  author  = {{OpenAI}},
  title   = {GPT-4o Mini: Advancing Cost-Efficient Intelligence},
  year    = {2024},
  url     = {https://openai.com/index/gpt-4o-mini-advancing-cost-efficient-intelligence/},
  note    = {Accessed: 2026-03-12}
}

@online{spiceworks_openai_structured_outputs_2024,
  author = {Anuj Mudaliar},
  title  = {OpenAI Launches Structured Outputs JSON API and Reduces GPT Prices},
  year   = {2024},
  url    = {https://www.spiceworks.com/tech/artificial-intelligence/news/openai-launches-structured-outputs-json-api-reduces-gpt-prices/},
  note   = {Accessed: 2026-03-12}
}

@online{towardsagi_o1_pricing_2024,
  author  = {Amdah H},
  title   = {OpenAI o1 API Pricing Explained: Everything You Need to Know},
  year    = {2024},
  url     = {https://medium.com/towards-agi/openai-o1-api-pricing-explained-everything-you-need-to-know-cbab89e5200d},
  note    = {Medium (Towards AGI). Accessed: 2026-03-12}
}

@online{openai_devs_twitter_pricing,
  author  = {{OpenAI Developers}},
  title   = {Announcement of GPT-4 Turbo API price reduction},
  year    = {2024},
  url     = {https://x.com/OpenAIDevs/status/1851668229938159853},
  note    = {Post on X (Twitter). Accessed: 2026-03-12}
}

@online{openai_realtime_api_updates_2024,
  author  = {Jeff Shariss},
  title   = {Realtime API Updates: WebRTC, Cheaper Prices, 4o-mini, and More},
  year    = {2024},
  url     = {https://community.openai.com/t/realtime-api-updates-webrtc-cheaper-prices-4o-mini-and-more/1059962},
  note    = {OpenAI Developer Community. Accessed: 2026-03-12}
}

@online{trustpilot_verified,
  author       = {{Trustpilot}},
  title        = {Why are some reviews marked "Verified"?},
  year         = {2026},
  url          = {https://help.trustpilot.com/s/article/Why-are-some-reviews-marked-Verified?language=en_US},
  note         = {Accessed: 2026-03-17}
}

@online{verge_chatgpt_api_2023,
  author = {Clark, Mitchell},
  title = {OpenAI announces an API for ChatGPT and its Whisper speech-to-text tech},
  journal = {The Verge},
  year = {2023},
  month = {March},
  day = {1},
  url = {https://www.theverge.com/2023/3/1/23620783/chatgpt-api-openai-pricing-whisper}
}

@article{cruz_gonzalez_2017,
  title={Bias corrections for probit and logit models with two-way fixed effects},
  author={Cruz-Gonzalez, Mario and Fern{\'a}ndez-Val, Iv{\'a}n and Weidner, Martin},
  journal={The Stata Journal},
  volume={17},
  number={3},
  pages={517--545},
  year={2017},
  publisher={SAGE Publications Sage CA: Los Angeles, CA}
}

@article{hahn_kuersteiner_2011,
  title={Bias reduction for dynamic nonlinear panel models with fixed effects},
  author={Hahn, Jinyong and Kuersteiner, Guido},
  journal={Econometric Theory},
  volume={27},
  number={6},
  pages={1152--1191},
  year={2011},
  publisher={Cambridge University Press}
}

@online{enterpriseappstoday_openai_stats,
  author = {Elad, Barry},
  title = {OpenAI Statistics 2024: Revenue, Growth, Users and Facts},
  year = {2024},
  url = {https://www.enterpriseappstoday.com/stats/openai-statistics.html},
  note = {Accessed: 2026-03-29}
}

@online{firstpagesage_market_share,
  author    = {Evan Bailyn},
  title     = {Top Generative AI Chatbots by Market Share – April 2026},
  year      = {2026},
  url       = {https://firstpagesage.com/reports/top-generative-ai-chatbots/},
  note      = {First Page Sage. Accessed: 2026-03-29}
}

@online{softwareoasis_chatgpt_dominance,
  author    = {Michael Bernzweig},
  title     = {ChatGPT Dominance Data + Statistics},
  year      = {2025},
  url       = {https://softwareoasis.com/chatgpt-dominance-2/},
  note      = {Software Oasis. Accessed: 2026-03-30}
}

@online{feedough_openai_stats,
  author    = {Aashish Pahwa},
  title     = {100+ OpenAI Statistics 2026: Valuation, Revenue \& Market Share},
  year      = {2026},
  url       = {https://www.feedough.com/openai-statistics/},
  note      = {Feedough. Accessed: 2026-03-30}
}

@online{axios_chatgpt_users,
  author    = {Ina Fried},
  title     = {OpenAI says ChatGPT usage has doubled since last year},
  year      = {2024},
  url       = {https://www.axios.com/2024/08/29/openai-chatgpt-200-million-weekly-active-users},
  note      = {Axios. Accessed: 2026-03-30}
}

@book{wooldridge_2010,
  title={Econometric analysis of cross section and panel data},
  author={Wooldridge, Jeffrey M},
  year={2010},
  publisher={MIT press}
}

@article{brown2020language,
  title={Language models are few-shot learners},
  author={Brown, Tom and Mann, Benjamin and Ryder, Nick and Subbiah, Melanie and Kaplan, Jared D and Dhariwal, Prafulla and Neelakantan, Arvind and Shyam, Pranav and Sastry, Girish and Askell, Amanda and others},
  journal={Advances in neural information processing systems},
  volume={33},
  pages={1877--1901},
  year={2020}
}

@article{bommasani2021opportunities,
  title={On the opportunities and risks of foundation models},
  author={Bommasani, Rishi and Hudson, Drew A and Adeli, Ehsan and Altman, Russ and Arora, Simran and von Arx, Sydney and Bernstein, Michael S and Bohg, Jeannette and Bosselut, Antoine and Brunskill, Emma and others},
  journal={arXiv preprint arXiv:2108.07258},
  year={2021}
}

@inproceedings{clark2021all,
  title={All that’s ‘human’is not gold: Evaluating human evaluation of generated text},
  author={Clark, Elizabeth and August, Tal and Serrano, Sofia and Haduong, Nikita and Gururangan, Suchin and Smith, Noah A},
  booktitle={Proceedings of the 59th Annual Meeting of the Association for Computational Linguistics and the 11th International Joint Conference on Natural Language Processing (Volume 1: Long Papers)},
  pages={7282--7296},
  year={2021}
}

@article{kreps2022all,
  title={All the news that’s fit to fabricate: AI-generated text as a tool of media misinformation},
  author={Kreps, Sarah and McCain, R Miles and Brundage, Miles},
  journal={Journal of experimental political science},
  volume={9},
  number={1},
  pages={104--117},
  year={2022},
  publisher={Cambridge University Press}
}

@article{ferrara2016rise,
  title={The rise of social bots},
  author={Ferrara, Emilio and Varol, Onur and Davis, Clayton and Menczer, Filippo and Flammini, Alessandro},
  journal={Communications of the ACM},
  volume={59},
  number={7},
  pages={96--104},
  year={2016},
  publisher={ACM New York, NY, USA}
}

@article{muzumdar2025dead,
  title={The dead internet theory: a survey on artificial interactions and the future of social media},
  author={Muzumdar, Prathamesh and Cheemalapati, Sumanth and RamiReddy, Srikanth Reddy and Singh, Kuldeep and Kurian, George and Muley, Apoorva},
  journal={arXiv preprint arXiv:2502.00007},
  year={2025}
}

@article{walter2025artificial,
  title={Artificial influencers and the dead internet theory},
  author={Walter, Yoshija},
  journal={AI \& SOCIETY},
  volume={40},
  number={1},
  pages={239--240},
  year={2025},
  publisher={Springer}
}

@article{allcott2017social,
  title={Social media and fake news in the 2016 election},
  author={Allcott, Hunt and Gentzkow, Matthew},
  journal={Journal of economic perspectives},
  volume={31},
  number={2},
  pages={211--236},
  year={2017},
  publisher={American Economic Association 2014 Broadway, Suite 305, Nashville, TN 37203-2418}
}

@article{shin2026ai,
  title={AI in the Age of Fake (Imagined) Content},
  author={Shin, Jieun},
  journal={Available at SSRN 6353778},
  year={2026}
}

@article{park2024rise,
  title={The rise of generative artificial intelligence and the threat of fake news and disinformation online: Perspectives from sexual medicine},
  author={Park, Hyun Jun},
  journal={Investigative and Clinical Urology},
  volume={65},
  number={3},
  pages={199},
  year={2024}
}

@article{marcellino2023rise,
  title={The rise of generative AI and the coming era of social media manipulation 3.0: Next-generation Chinese astroturfing and coping with ubiquitous AI},
  author={Marcellino, William and Beauchamp-Mustafaga, Nathan and Kerrigan, Amanda and Chao, Lev N and Smith, Jackson},
  year={2023}
}

@article{rachmiani2024impact,
  title={The impact of online reviews and ratings on consumer purchasing decisions on e-commerce platforms},
  author={Rachmiani, R and Oktadinna, N Kintan and Fauzan, T Rachmat},
  journal={International Journal of Management Science and Information Technology},
  volume={4},
  number={2},
  pages={504--515},
  year={2024}
}

@article{lackermair2013importance,
  title={Importance of online product reviews from a consumer’s perspective},
  author={Lackermair, Georg and Kailer, Daniel and Kanmaz, Kenan},
  journal={Advances in economics and business},
  volume={1},
  number={1},
  pages={1--5},
  year={2013}
}

@misc{qiu2023online,
  title={How online reviews affect purchase intention: A meta-analysis across contextual and cultural factors. Data and Information Management, 8 (2), 100058},
  author={Qiu, K and Zhang, L},
  year={2023}
}

@online{burton2024reviews,
  author = {Burton, Joe},
  title = {Online Reviews Can Make Or Break Your Business: Pay Attention To Them},
  year = {2024},
  month = {November},
  day = {12},
  url = {https://www.forbes.com/councils/forbestechcouncil/2024/11/12/online-reviews-can-make-or-break-your-business-pay-attention-to-them/},
  note = {Forbes Technology Council}
}

@article{lim2025rise,
  title={The rise of fake reviews: Toward a marketing-oriented framework for understanding fake reviews},
  author={Lim, Weng Marc and Agarwal, Reeti and Mishra, Anubhav and Mehrotra, Ankit},
  journal={Australasian Marketing Journal},
  volume={33},
  number={2},
  pages={178--198},
  year={2025},
  publisher={SAGE Publications Sage UK: London, England}
}

@article{martinez2021fake,
  title={Fake reviews on online platforms: perspectives from the US, UK and EU legislations},
  author={Mart{\'\i}nez Otero, Juan Mar{\'\i}a},
  journal={SN Social Sciences},
  volume={1},
  number={7},
  pages={181},
  year={2021},
  publisher={Springer}
}

@article{he2022market,
  title={The market for fake reviews},
  author={He, Sherry and Hollenbeck, Brett and Proserpio, Davide},
  journal={Marketing Science},
  volume={41},
  number={5},
  pages={896--921},
  year={2022},
  publisher={INFORMS}
}

@inbook{inbook,
author = {Özaydın, Haydar},
year = {2025},
month = {03},
pages = {},
title = {Fake Reviews and Ratings Undermining Consumer Trust},
isbn = {978-625-5958-72-3},
doi = {10.58830/ozgur.pub710.c3028}
}

@article{gupta2024recent,
  title={Recent state-of-the-art of fake review detection: a comprehensive review},
  author={Gupta, Richa and Jindal, Vinita and Kashyap, Indu},
  journal={The Knowledge Engineering Review},
  volume={39},
  pages={e8},
  year={2024},
  publisher={Cambridge University Press}
}

@article{zhao2025ai,
  title={AI vs. human: A large-scale analysis of AI-generated fake reviews, human-generated fake reviews and authentic reviews},
  author={Zhao, Yuexin and Tang, Siyi and Zhang, Hongyu and Lyu, Long},
  journal={Journal of Retailing and Consumer Services},
  volume={87},
  pages={104400},
  year={2025},
  publisher={Elsevier}
}

@article{meng2025large,
  title={Large Language Models as' Hidden Persuaders': Fake Product Reviews are Indistinguishable to Humans and Machines},
  author={Meng, Weiyao and Harvey, John and Goulding, James and Carter, Chris James and Lukinova, Evgeniya and Smith, Andrew and Frobisher, Paul and Forrest, Mina and Nica-Avram, Georgiana},
  journal={arXiv preprint arXiv:2506.13313},
  year={2025}
}

@article{knight2023generative,
  title={Generative AI and the Perceived Quality of User-Generated Content: Evidence from Online Reviews},
  author={Knight, Samsun and Bart, Yakov and Yang, Minwen},
  journal={Northeastern U. D’Amore-McKim School of Business Research Paper},
  number={4621982},
  year={2023}
}

@article{gambetti2023dissecting,
  title={Dissecting ai-generated fake reviews: detection and analysis of GPT-based restaurant reviews on social media},
  author={Gambetti, Alessandro and Han, Qiwei},
  year={2023}
}

@article{luo2026ai,
  title={AI-generated fake review detection},
  author={Luo, Jiwei and Nan, Guofang and Li, Dahui},
  journal={Decision Support Systems},
  pages={114628},
  year={2026},
  publisher={Elsevier}
}

@article{gambetti2023combat,
  title={Combat ai with ai: Counteract machine-generated fake restaurant reviews on social media},
  author={Gambetti, Alessandro and Han, Qiwei},
  journal={arXiv preprint arXiv:2302.07731},
  year={2023}
}

@article{pocchiari2025online,
  title={Online reviews: A literature review and roadmap for future research},
  author={Pocchiari, Martina and Proserpio, Davide and Dover, Yaniv},
  journal={International journal of research in marketing},
  volume={42},
  number={2},
  pages={275--297},
  year={2025},
  publisher={Elsevier}
}

@article{alzate2021online,
  title={Online reviews and product sales: the role of review visibility},
  author={Alzate, Miriam and Arce-Urriza, Marta and Cebollada, Javier},
  journal={Journal of Theoretical and Applied Electronic Commerce Research},
  volume={16},
  number={4},
  pages={638--669},
  year={2021},
  publisher={MDPI}
}

@article{huang2020impact,
  title={The impact of online consumer reviews on online sales: the case-based decision theory approach},
  author={Huang, M and Pape, AD},
  journal={Journal of Consumer Policy},
  volume={43},
  number={3},
  pages={463--490},
  year={2020},
  publisher={Springer}
}

@article{mayzlin2014promotional,
  title={Promotional reviews: An empirical investigation of online review manipulation},
  author={Mayzlin, Dina and Dover, Yaniv and Chevalier, Judith},
  journal={American Economic Review},
  volume={104},
  number={8},
  pages={2421--2455},
  year={2014},
  publisher={American Economic Association 2014 Broadway, Suite 305, Nashville, TN 37203}
}

@article{luca2016fake,
  title={Fake it till you make it: Reputation, competition, and Yelp review fraud},
  author={Luca, Michael and Zervas, Georgios},
  journal={Management science},
  volume={62},
  number={12},
  pages={3412--3427},
  year={2016},
  publisher={INFORMS}
}

@online{trustpilot_trust,
  author       = {{Trustpilot}},
  title        = {Trust and Transparency},
  year         = {2026},
  howpublished = {\url{https://corporate.trustpilot.com/trust}},
  note         = {Accessed: 2026-04-21}
}

@article{raman2024exploring,
  title={Exploring university students’ adoption of ChatGPT using the diffusion of innovation theory and sentiment analysis with gender dimension},
  author={Raman, Raghu and Mandal, Santanu and Das, Payel and Kaur, Tavleen and Sanjanasri, JP and Nedungadi, Prema},
  journal={Human Behavior and Emerging Technologies},
  volume={2024},
  number={1},
  pages={3085910},
  year={2024},
  publisher={Wiley Online Library}
}

@article{zhang2024exploring,
  title={Exploring the latest applications of openai and ChatGPT: An in-depth survey.},
  author={Zhang, Hong and Shao, Haijian},
  journal={Computer Modeling in Engineering \& Sciences (CMES)},
  volume={138},
  number={3},
  year={2024}
}

@article{hu2009online,
  title={Why do online product reviews have a J-shaped distribution? Overcoming biases in online word-of-mouth communication},
  author={Hu, Nan and Pavlou, Paul A and Zhang, Jie Jennifer},
  journal={Communications of the ACM},
  volume={52},
  number={10},
  pages={144--147},
  year={2009}
}

@article{chevalier2006effect,
  title={The effect of word of mouth on sales: Online book reviews},
  author={Chevalier, Judith A and Mayzlin, Dina},
  journal={Journal of marketing research},
  volume={43},
  number={3},
  pages={345--354},
  year={2006},
  publisher={SAGE Publications Sage CA: Los Angeles, CA}
}

@article{tanase2024online,
  title={Online Reviews in Romania: Motivations, Perceptions, and the Impact of the J-Shaped Distribution on Consumer Behavior},
  author={Tanase, Ionut and Barbu, Lucia Nicoleta and Grejdan, Elena Florentina},
  journal={Ovidius University Annals, Economic Sciences Series},
  volume={24},
  number={2},
  pages={450--454},
  year={2024},
  publisher={Ovidius University of Constantza, Faculty of Economic Sciences}
}

@article{kujur2025comparative,
  title={A Comparative Analysis of AI-Generated and Human-Written Text: Linguistic Patterns, Detection Accuracy, and Implications for Modern Communication},
  author={Kujur, Arahan},
  journal={Detection Accuracy, and Implications for Modern Communication (November 29, 2025)},
  year={2025}
}

@article{culda2025comparative,
  title={Comparative linguistic analysis framework of human-written vs. machine-generated text},
  author={Culda, Lia Cornelia and Neri{\c{s}}anu, Raluca Andreea and Cristescu, Marian Pompiliu and Mara, Dumitru Alexandru and B{\^a}ra, Adela and Oprea, Simona-Vasilica},
  journal={Connection Science},
  volume={37},
  number={1},
  pages={2507183},
  year={2025},
  publisher={Taylor \& Francis}
}

@article{munoz2024contrasting,
  title={Contrasting linguistic patterns in human and LLM-generated news text},
  author={Mu{\~n}oz-Ortiz, Alberto and G{\'o}mez-Rodr{\'\i}guez, Carlos and Vilares, David},
  journal={Artificial Intelligence Review},
  volume={57},
  number={10},
  pages={265},
  year={2024},
  publisher={Springer}
}

@article{fariello2024distinguishing,
  title={Distinguishing human from machine: A review of advances and challenges in ai-generated text detection},
  author={Fariello, Serena},
  year={2024},
  publisher={UNIR}
}

@article{guo2024detective,
  title={Detective: Detecting ai-generated text via multi-level contrastive learning},
  author={Guo, Xun and Zhang, Shan and He, Yongxin and Zhang, Ting and Feng, Wanquan and Huang, Haibin and Ma, Chongyang},
  journal={Advances in Neural Information Processing Systems},
  volume={37},
  pages={88320--88347},
  year={2024}
}

@article{chaka2024reviewing,
  title={Reviewing the performance of AI detection tools in differentiating between AI-generated and human-written texts: A literature and integrative hybrid review},
  author={Chaka, Chaka},
  journal={Journal of Applied Learning \& Teaching},
  volume={7},
  number={1},
  pages={115--126},
  year={2024},
  publisher={Kaplan Business School Australia Sydney, NSW}
}

@online{baronio2025deadinternet,
  author       = {Joseph Baronio},
  title        = {Is the Internet Dead?},
  year         = {2025},
  month        = feb,
  day          = {5},
  organization = {ABC News Australia},
  url          = {https://www.abc.net.au/btn/high/is-the-internet-dead/104897518},
  urldate      = {2026-08-02},
  note         = {Behind the News (BTN)}
}

@online{down2025aislop,
  author       = {Aisha Down},
  title        = {From Shrimp Jesus to Erotic Tractors: How Viral AI Slop Took Over the Internet},
  year         = {2025},
  month        = dec,
  day          = {27},
  organization = {The Guardian},
  url          = {https://www.theguardian.com/technology/2025/dec/27/from-shrimp-jesus-to-erotic-tractors-how-viral-ai-slop-took-over-the-internet},
  urldate      = {2026-08-02}
}

@online{murray2025deadinternet,
  author       = {Conor Murray},
  title        = {Ohanian And Altman Warn Of {`Dead Internet Theory'}---What Is It And How Is AI Making It Happen?},
  year         = {2025},
  month        = oct,
  day          = {13},
  organization = {Forbes},
  url          = {https://www.forbes.com/sites/conormurray/2025/10/13/ohanian-and-altman-warn-of-dead-internet-theory-what-is-it-and-how-is-ai-making-it-happen/},
  urldate      = {2026-08-02}
}

@online{levy2026aislop,
  author       = {Steven Levy},
  title        = {AI Slop Melodramas Are Taking Over X---and Their Creators Are Cashing In},
  year         = {2026},
  month        = jul,
  day          = {31},
  organization = {WIRED},
  url          = {https://www.wired.com/story/ai-slop-melodramas-are-taking-over-x-and-their-creators-are-cashing-in/},
  urldate      = {2026-08-02}
}

@article{santos2025improving,
  title={Improving trust in online reviews: a machine learning approach to detecting artificial intelligence-generated reviews},
  author={Santos, Ana Marta and Antonio, Nuno},
  journal={Information Technology \& Tourism},
  volume={27},
  number={3},
  pages={739--766},
  year={2025},
  publisher={Springer}
}

@inproceedings{agrahari2025can,
  title={Can You Really Trust That Review? ProtoFewRoBERTa and DetectAIRev: A Prototypical Few-Shot Method and Multi-Domain Benchmark for Detecting AI-Generated Reviews},
  author={Agrahari, Shifali and Kumar, Sujit and Sanasam, Ranbir Singh},
  booktitle={Proceedings of the 14th International Joint Conference on Natural Language Processing and the 4th Conference of the Asia-Pacific Chapter of the Association for Computational Linguistics},
  pages={2118--2140},
  year={2025}
}

@techreport{tully2024state,
  author      = {Tully, Tim and Redfern, Joff and Xiao, Derek},
  title       = {2024: The State of Generative AI in the Enterprise},
  institution = {Menlo Ventures},
  year        = {2024},
  url         = {https://menlovc.com/2024-the-state-of-generative-ai-in-the-enterprise/},
  note        = {Accessed: 2026-08-30}
}

@misc{wang2024generative,
  author       = {Wang, Sarah and Xu, Shangda},
  title        = {16 Changes to the Way Enterprises Are Building and Buying Generative AI},
  year         = {2024},
  month        = {March},
  publisher    = {Andreessen Horowitz},
  url          = {https://a16z.com/generative-ai-enterprise-2024/},
  note         = {Accessed: 2026-08-30}
}

\clearpage

\section*{Supplementary Materials}\label{sec: supp}

\renewcommand{\thefigure}{S\arabic{figure}}
\setcounter{figure}{0}

\renewcommand{\thetable}{S\arabic{table}}
\setcounter{table}{0}

\begin{figure}[H]
    \centering

    \caption{OpenAI Model Pricing Reductions and Releases, 2023--2024 (raw prices)}
    \label{fig:price_raw}

    \begin{minipage}{0.8\textwidth}
        \centering
        \includegraphics[width=\textwidth]{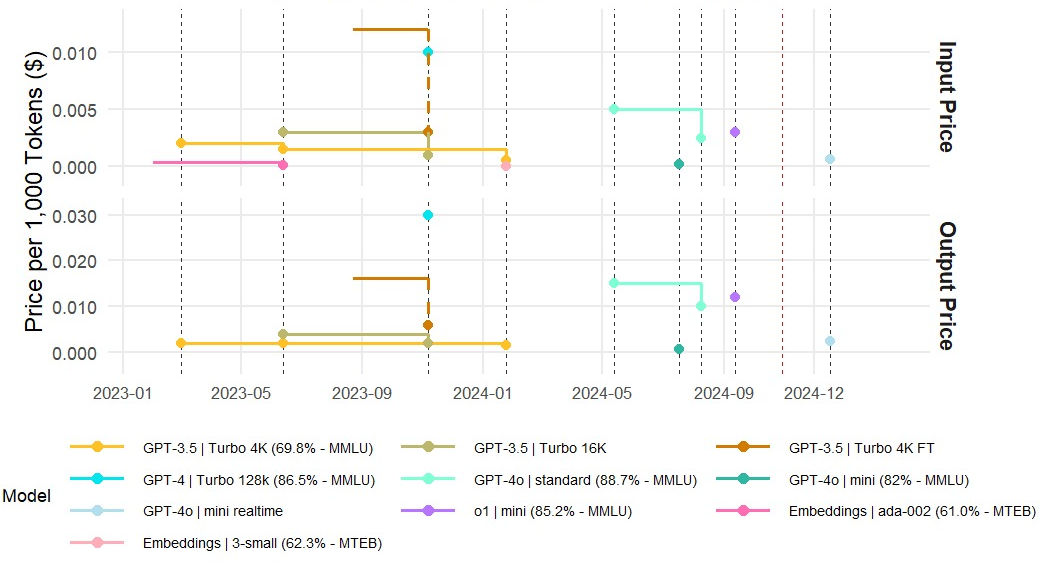}

        \vspace{0.3em}

        {\scriptsize
\justifying
        \noindent\textit{Notes:} This figure presents the raw (non-normalized) pricing series underlying Figure~\ref{fig:price_percent}. Prices are reported in U.S. dollars per 1{,}000 tokens. The upper panel shows input prices and the lower panel shows output prices. The horizontal axis displays calendar dates over the 2023--2024 period. Colors correspond to different models, with similar color schemes used for models belonging to the same family. Vertical dashed lines indicate the dates of the LLM supply shocks used in the empirical analysis and listed in Table~\ref{tab:openai_price_reductions}. Solid circular markers denote observed pricing points included in the analysis, corresponding either to price reductions or to the release of lower-priced models. Embedding models do not report output prices because they generate vector representations rather than textual outputs. Figure~\ref{fig:price_percent} presents the same data normalized to each model's initial observed price (=100\%).
        \par}
    \end{minipage}

\end{figure}

\setcounter{section}{0}
\renewcommand{\thesection}{S\arabic{section}}

\section{Event-study normalization and derivation}
\label{sec:supp_eventstudy}

This section provides the derivation of the event-study specification used in the main analysis. The specification normalizes the average of the seven pre-event coefficients to zero rather than selecting a single pre-event day as the omitted reference period. The normalization is imposed separately on the event-time coefficients, $\lambda_k$, and the treatment-by-event-time interaction coefficients, $\delta_k$.

Each observation is indexed by company $c$, verification status $v$, and calendar date $t$. Define

\[
\mathrm{Treatment}_v
=
\mathbf{1}\{v=\text{unverified}\},
\]

such that $\mathrm{Treatment}_v=1$ for unverified observations and $\mathrm{Treatment}_v=0$ for verified observations.

Let

\[
D_t^k
=
\mathbf{1}\{\text{date } t \text{ is } k \text{ days relative to the event date}\}.
\]

The event-time indicator is indexed by calendar date $t$ and relative time $k$, where $k$ denotes the number of days relative to the event date. For a given calendar date $t$, the verified and unverified observations therefore have the same event-time indicator. Differences between the two groups are captured by $\mathrm{Treatment}_v$ and its interactions with the event-time indicators. 

The event day, $k=0$, is excluded from the specification. The event-study window consists of the seven days preceding and seven days following the event:

\[
k \in \{-7,-6,\ldots,-1,1,\ldots,7\}.
\]

\subsection{Normalization of the pre-event event-time coefficients}

Rather than setting one particular pre-event coefficient equal to zero, we normalize the event-time coefficients such that their average over the seven pre-event days equals zero:

\[
\frac{1}{7}\sum_{k=-7}^{-1}\lambda_k=0.
\]

Equivalently,

\[
\sum_{k=-7}^{-1}\lambda_k=0.
\]

Solving this restriction for the coefficient on relative day $-1$ gives

\[
\lambda_{-1}
=
-\sum_{k=-7}^{-2}\lambda_k.
\]

Without imposing the normalization, the pre-event event-time component can be written as

\[
\sum_{k=-7}^{-1}\lambda_k D_t^k.
\]

Separating the coefficient for relative day $-1$ yields

\[
\sum_{k=-7}^{-2}\lambda_k D_t^k
+
\lambda_{-1}D_t^{-1}.
\]

Substituting

\[
\lambda_{-1}
=
-\sum_{k=-7}^{-2}\lambda_k
\]

gives

\[
\sum_{k=-7}^{-2}\lambda_k D_t^k
-
\left(
\sum_{k=-7}^{-2}\lambda_k
\right)
D_t^{-1}.
\]

Collecting the terms associated with each $\lambda_k$, the expression becomes

\[
\sum_{k=-7}^{-2}
\lambda_k
\left(
D_t^k-D_t^{-1}
\right).
\]

Thus, using $D_t^k-D_t^{-1}$ as the regressors for $k=-7,\ldots,-2$ is algebraically equivalent to estimating all seven pre-event event-time coefficients subject to the restriction that their average equals zero.

\subsection{Normalization of the pre-event interaction coefficients}

We impose the analogous normalization on the treatment-by-event-time interaction coefficients:

\[
\frac{1}{7}\sum_{k=-7}^{-1}\delta_k=0.
\]

Equivalently,

\[
\sum_{k=-7}^{-1}\delta_k=0.
\]

Solving for the interaction coefficient on relative day $-1$ gives

\[
\delta_{-1}
=
-\sum_{k=-7}^{-2}\delta_k.
\]

Without imposing the normalization, the pre-event interaction component can be written as

\[
\sum_{k=-7}^{-1}
\delta_k
\left(
\mathrm{Treatment}_v D_t^k
\right).
\]

Separating the interaction coefficient for relative day $-1$ yields

\[
\sum_{k=-7}^{-2}
\delta_k
\left(
\mathrm{Treatment}_v D_t^k
\right)
+
\delta_{-1}
\left(
\mathrm{Treatment}_v D_t^{-1}
\right).
\]

Substituting

\[
\delta_{-1}
=
-\sum_{k=-7}^{-2}\delta_k
\]

gives

\[
\sum_{k=-7}^{-2}
\delta_k
\left(
\mathrm{Treatment}_v D_t^k
\right)
-
\left(
\sum_{k=-7}^{-2}\delta_k
\right)
\left(
\mathrm{Treatment}_v D_t^{-1}
\right).
\]

Collecting the terms associated with each $\delta_k$, this expression becomes

\[
\sum_{k=-7}^{-2}
\delta_k
\mathrm{Treatment}_v
\left(
D_t^k-D_t^{-1}
\right).
\]

Thus, using
$\mathrm{Treatment}_v(D_t^k-D_t^{-1})$
as the interaction regressors for $k=-7,\ldots,-2$ is algebraically equivalent to estimating all seven pre-event interaction coefficients subject to the restriction that their average equals zero.

\subsection{Resulting regression specification}
\label{subsec:regression_specification}

Combining these reparameterizations with the post-event coefficients gives the event-study regression

\[
\begin{aligned}
Y_{cvt}
={}&
\alpha_c
+
\gamma_{w(t)}
+
\theta\,\mathrm{Treatment}_v
\\
&+
\sum_{k=-7}^{-2}
\lambda_k
\left(
D_t^k-D_t^{-1}
\right)
+
\sum_{k=1}^{7}
\lambda_k D_t^k
\\
&+
\sum_{k=-7}^{-2}
\delta_k
\mathrm{Treatment}_v
\left(
D_t^k-D_t^{-1}
\right)
+
\sum_{k=1}^{7}
\delta_k
\mathrm{Treatment}_v D_t^k
+
\varepsilon_{cvt}.
\end{aligned}
\]

\begin{table}[H]
\centering
\caption{Difference-in-Differences Results: Review Count, Length, and Text Homogeneity}
\label{tab:did_nonsig_explanatory_main}
\begin{tabular}{lccccc}
\hline\hline
 & (1) & (2) & (3) & (4) & (5) \\
 & Count & Log count & Length & Log length & Homogeneity \\
\hline
Treatment 
& -0.066*** & 0.012*** & 34.841*** & 0.670*** & 0.0363*** \\
& (0.013) & (0.001) & (0.591) & (0.010) & (0.0015) \\

After 
& -0.023*** & -0.008*** & 0.269 & -0.007* & 0.0006 \\
& (0.003) & (0.000) & (0.213) & (0.004) & (0.0006) \\

Treatment $\times$ After 
& -0.004 & -0.001*** & 0.227 & 0.003 & -0.0009 \\
& (0.004) & (0.000) & (0.215) & (0.004) & (0.0006) \\

\hline
Company FE & Yes & Yes & Yes & Yes & Yes \\
Week FE & Yes & Yes & Yes & Yes & Yes \\
Mean (After = 0) & 0.176 & 0.053 & 47.902 & 3.421 & 0.604 \\
R-squared & 0.178 & 0.355 & 0.297 & 0.392 & 0.344 \\
Observations & 16,116,186 & 16,116,186 & 748,513 & 748,513 & 1,209,426 \\
\hline\hline
\end{tabular}

\vspace{0.5em}
\begin{minipage}{0.95\textwidth}
\footnotesize
\textit{Notes:} This table reports Difference-in-Differences estimates for additional standard review metrics, including review count, the logarithm of review count, review length, the logarithm of review length, and review text homogeneity. Across most specifications, the coefficient on the interaction term \\(\textit{Treatment} $\times$ \textit{After}) is small and not statistically significant, indicating no robust differential effect of the LLM supply shocks on these outcomes. An exception is the specification using the logarithm of review counts, where the interaction coefficient is statistically significant in the baseline model. However, this result is not robust: in the corresponding event-study specification, the parallel trends assumption is violated, and the effect does not persist (see Table \ref{tab:pre_post_tests_main}).

Standard errors in parentheses. Statistical significance levels: *** $p < 0.01$, ** $p < 0.05$, * $p < 0.1$.
\end{minipage}
\end{table}

% =========================
% Table 2
% =========================

\begin{table}[H]
\centering
\caption{Difference-in-Differences Results: Log Review Counts by Rating Category (1--5)}
\label{tab:did_nonsig_log_counts}

\begin{threeparttable}

\begin{tabular}{lccccc}
\toprule
 & (1) & (2) & (3) & (4) & (5) \\
 & Log count 1 & Log count 2 & Log count 3 & Log count 4 & Log count 5 \\
\midrule

Treatment
& 0.01636*** & -0.00037*** & -0.00181*** & -0.00339*** & -0.00576*** \\
& (0.00036) & (0.00013) & (0.00018) & (0.00028) & (0.00094) \\

After
& -0.00217*** & -0.00033*** & -0.00033*** & -0.00079*** & -0.00582*** \\
& (0.00008) & (0.00004) & (0.00005) & (0.00007) & (0.00017) \\

Treatment $\times$ After
& -0.00035*** & 0.00008** & 0.00003 & 0.00003 & -0.00038** \\
& (0.00009) & (0.00004) & (0.00004) & (0.00006) & (0.00015) \\

\midrule
Company FE & Yes & Yes & Yes & Yes & Yes \\
Week FE & Yes & Yes & Yes & Yes & Yes \\
Mean (After = 0) & 0.014 & 0.002 & 0.003 & 0.006 & 0.038 \\
R-squared & 0.239 & 0.167 & 0.207 & 0.232 & 0.332 \\
Observations & 16,116,186 & 16,116,186 & 16,116,186 & 16,116,186 & 16,116,186 \\
\bottomrule
\end{tabular}

\begin{tablenotes}[flushleft]
\footnotesize
\item \textit{Notes:} This table reports Difference-in-Differences estimates for the logarithm of review counts across rating categories (1 to 5 stars). The interaction term (\textit{Treatment} $\times$ \textit{After}) is statistically significant for 1-star, 2-star, and 5-star reviews in the baseline specification, while no significant effects are observed for 3-star and 4-star reviews. However, these results are not robust: in the corresponding event-study analysis, the parallel trends assumption is violated for these outcomes, and the estimated effects do not persist (see Table~\ref{tab:pre_post_tests_main}).

Standard errors are reported in parentheses. \\Statistical significance levels are denoted as follows: *** $p<0.01$, ** $p<0.05$, * $p<0.1$.
\end{tablenotes}

\end{threeparttable}
\end{table}

\begin{table}[H]
\centering
\caption{Pre-Trend and Post-Treatment Tests}
\label{tab:pre_post_tests_main}

\resizebox{\textwidth}{!}{
\begin{tabular}{lccc}
\toprule
Outcome 
& Pre-trend p-value 
& Post-treatment p-value 
& Average post coefficient \\
\midrule

Count
& 0.007***
& 0.114
& -0.004255 \\

Log count
& $< 0.001$***
& $< 0.001$***
& -0.000721 \\

Length
& 0.640
& 0.089*
& 0.255823 \\

Log length
& 0.599
& 0.122
& 0.003607 \\

Homogeneity
& 0.270
& 0.285
& -0.000647 \\

Log count 1
& $< 0.001$***
& $< 0.001$***
& -0.000349 \\

Log count 2
& 0.022**
& 0.273
& 0.000081 \\

Log count 3
& 0.590
& 0.942
& 0.000025 \\

Log count 4
& 0.089*
& 0.030**
& 0.000031 \\

Log count 5
& $< 0.001$***
& $< 0.001$***
& -0.000377 \\

\bottomrule
\end{tabular}
}

\vspace{0.15cm}

\begin{minipage}{\textwidth}
\footnotesize
\textit{Notes:} 
This table reports joint F-tests of pre-treatment and post-treatment interaction coefficients from the event-study specifications. 
The pre-treatment test examines the null hypothesis that the freely estimated pre-treatment interaction coefficients are jointly equal to zero. 
Treatment-by-event-time coefficients are normalized such that their average over the seven pre-treatment days ($k=-7,\ldots,-1$) equals zero; the coefficient at $k=-1$ is recovered from this restriction. 
The post-treatment test examines the null hypothesis that all post-treatment interaction coefficients ($k=1,\ldots,7$) are jointly equal to zero. 
The average post coefficient is the arithmetic mean of the seven post-treatment interaction coefficients, 
$\frac{1}{7}\sum_{k=1}^{7}\delta_k$.
All regressions include company and week fixed effects, and standard errors are clustered at the company level.
Statistical significance levels are denoted as follows: *** $p<0.01$, ** $p<0.05$, * $p<0.1$.
\end{minipage}
\end{table}

\begin{figure}[H]
\centering

\caption{Event-Study Coefficients for Total Count, Length, and Homogeneity}
\label{fig:event_study_main_homog}

\begin{minipage}{0.95\textwidth}
\centering
\includegraphics[width=\textwidth]{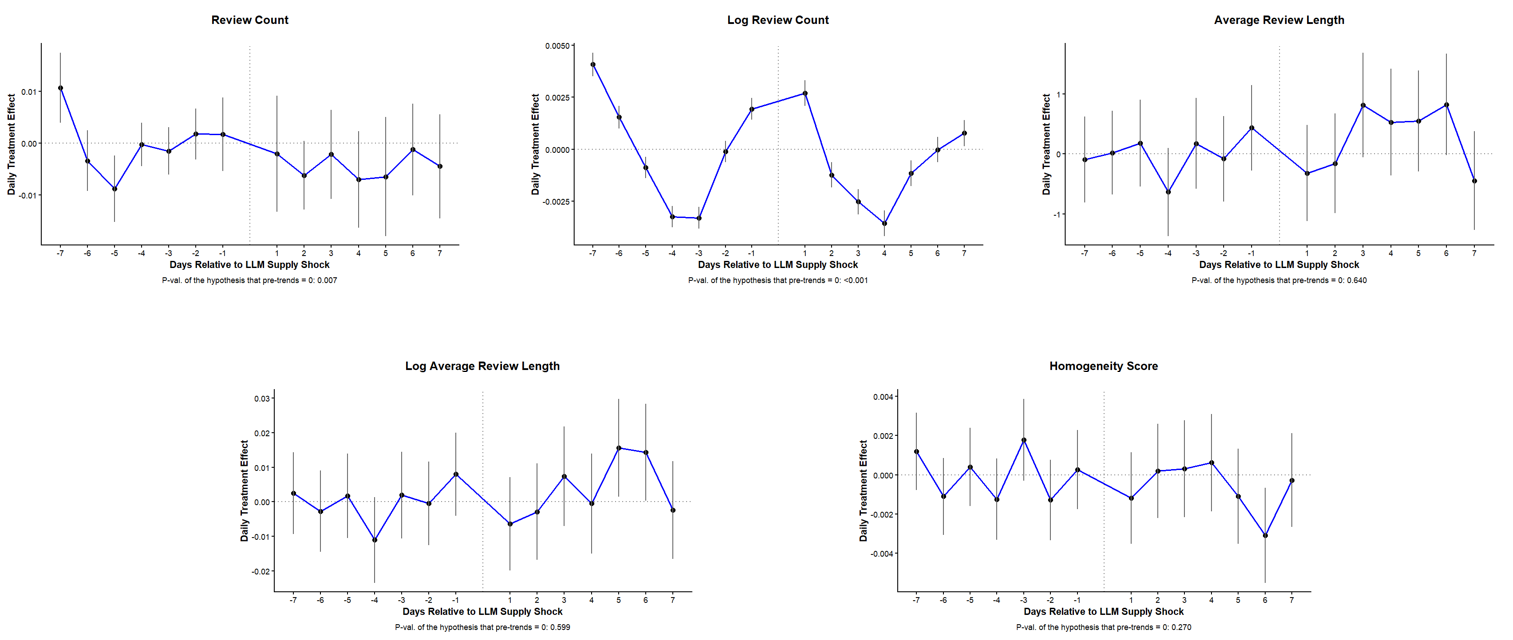}

\vspace{0.3em}

{\scriptsize
\justifying
\noindent
\textit{Notes:} This figure plots the event-study treatment-by-event-time coefficients for review count, log review count, average review length, log average review length, and text homogeneity. Points represent estimated treatment effects at each event time, and vertical bars show 95\% confidence intervals. The treatment effects are normalized such that the average coefficient over the seven pre-treatment days ((k=-7,\ldots,-1)) equals zero. To impose this normalization, the pre-treatment indicators are parameterized relative to (k=-1), and the coefficient for (k=-1) is recovered from the restriction that the seven pre-treatment coefficients sum to zero. The vertical line marks the LLM supply shock dates, and the horizontal dashed line denotes zero.
\par}

\end{minipage}

\end{figure}

\begin{figure}[H]
\centering

\caption{Event-Study Coefficients for Log Review Counts by Rating Category}
\label{fig:event_study_log_counts}

\begin{minipage}{0.95\textwidth}
\centering
\includegraphics[width=\textwidth]{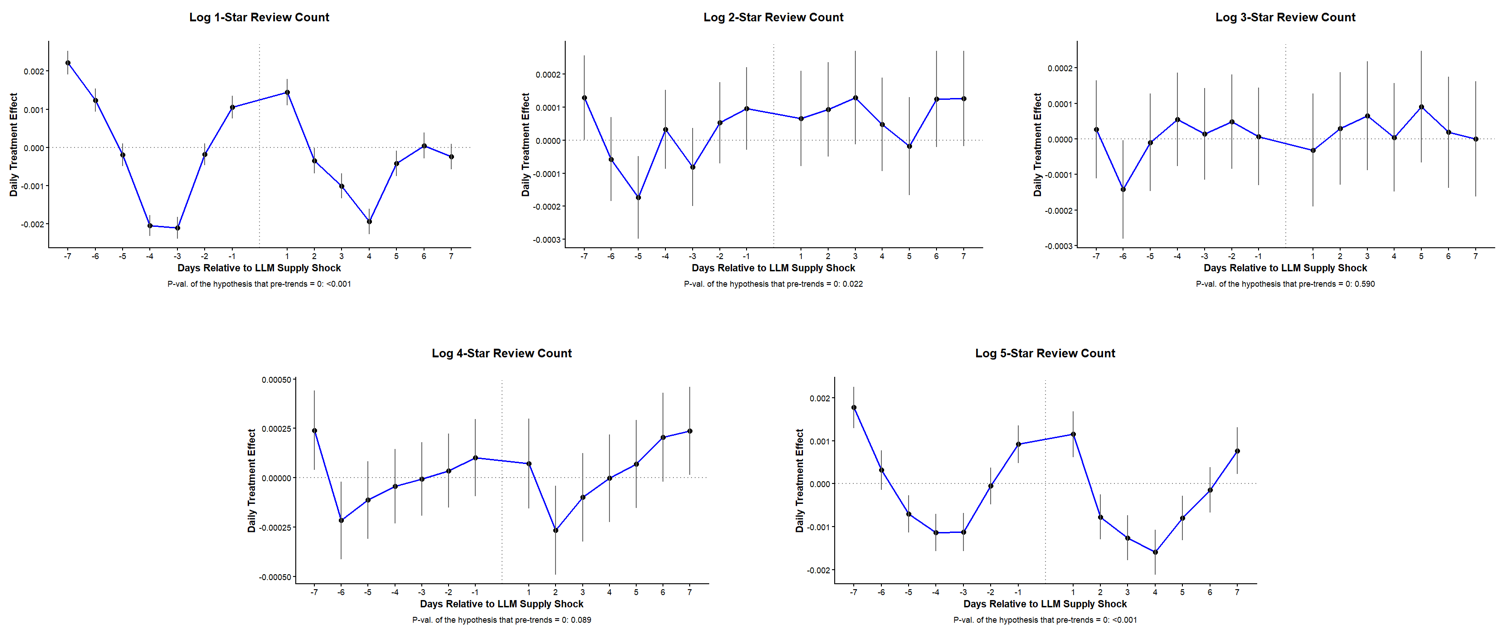}

\vspace{0.3em}

{\scriptsize
\justifying
\noindent
\textit{Notes:}This figure plots the event-study treatment-by-event-time coefficients for the logarithm of review counts by rating category, separately for 1-star through 5-star reviews. Points represent estimated treatment effects at each event time, and vertical bars show 95\% confidence intervals. The treatment effects are normalized such that the average coefficient over the seven pre-treatment days ((k=-7,\ldots,-1)) equals zero. To impose this normalization, the pre-treatment indicators are parameterized relative to (k=-1), and the coefficient for (k=-1) is recovered from the restriction that the seven pre-treatment coefficients sum to zero. The vertical line marks the LLM supply shock dates, and the horizontal dashed line denotes zero.
\par}

\end{minipage}

\end{figure}

\begin{table}[H]
\centering
\caption{Difference-in-Differences Results for Combined 1-Star and 5-Star Review-Volume Categories Using Adjusted Thresholds}
\label{tab:did_r1r5_adjusted}

\begin{tabular}{lccc}
\toprule
& (1) & (2) & (3) \\
& \multicolumn{1}{c}{1+5-Star Rating} 
& \multicolumn{1}{c}{1+5-Star Rating} 
& \multicolumn{1}{c}{1+5-Star Rating} \\
& \multicolumn{1}{c}{Adjusted Threshold}
& \multicolumn{1}{c}{Adjusted Threshold}
& \multicolumn{1}{c}{Adjusted Threshold} \\
& \multicolumn{1}{c}{Low-volume: 0--17}
& \multicolumn{1}{c}{Moderate-volume: 18--84}
& \multicolumn{1}{c}{High-volume: 85+} \\
\midrule

Treatment
& 0.001243*** 
& -0.001077*** 
& -0.000167*** \\

& (0.00013)
& (0.00012)
& (0.00004) \\

After
& 0.000224*** 
& -0.000217*** 
& -0.000007 \\

& (0.00003)
& (0.00003)
& (0.00001) \\

Treatment $\times$ After
& -0.000035
& 0.000036
& -0.000001 \\

& (0.00002)
& (0.00002)
& (0.00001) \\

\midrule

Company FE
& Yes
& Yes
& Yes \\

Week FE
& Yes
& Yes
& Yes \\

Mean (After $=0$)
& 0.9988
& 0.0011
& 0.0001 \\

R-squared
& 0.265
& 0.241
& 0.215 \\

Observations
& 16,116,186
& 16,116,186
& 16,116,186 \\

\bottomrule
\end{tabular}

\vspace{0.15cm}

\begin{minipage}{0.95\textwidth}
\footnotesize
\textit{Notes:} 
This table reports fixed effects difference-in-differences regressions for indicators capturing combined 1-star and 5-star review-volume categories using adjusted thresholds. The dependent variables are indicators equal to one if a company--day--verification-status observation belongs to the low-volume (0--17 combined 1-star and 5-star reviews), moderate-volume (18--84 combined 1-star and 5-star reviews), or high-volume (85 or more combined 1-star and 5-star reviews) category, and zero otherwise. The coefficients can therefore be interpreted as changes in the probability that an observation belongs to the corresponding review-volume category. Treatment equals one for unverified reviews and zero for verified reviews. After equals one for observations after the LLM supply shock and zero for observations before the shock. The coefficient on Treatment $\times$ After is the difference-in-differences estimate. All specifications include company and week fixed effects, and standard errors are clustered at the company level. Standard errors are reported in parentheses. Statistical significance levels are denoted as follows: *** $p<0.01$, ** $p<0.05$, * $p<0.10$.
\end{minipage}

\end{table}

\begin{figure}[H]
    \centering
\caption{Event-Study Estimates for Combined 1-Star and 5-Star Review-Volume Categories Using Adjusted Thresholds}
    \includegraphics[width=\textwidth]{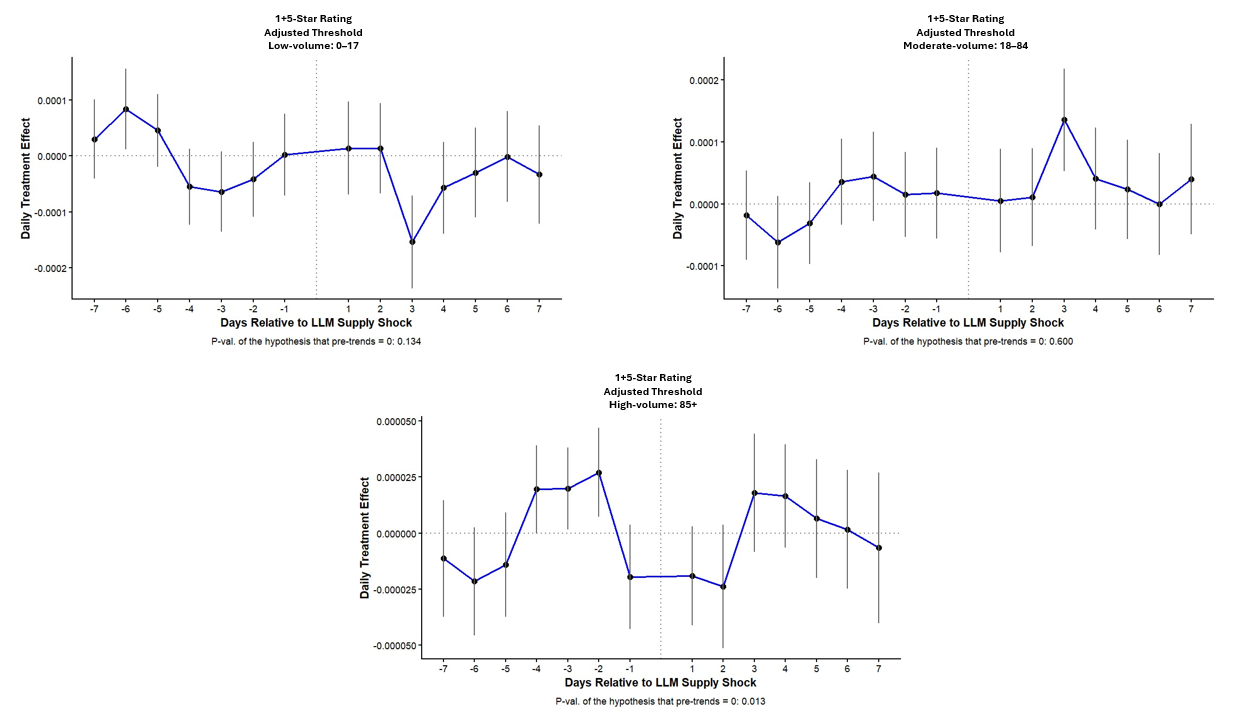}

    \label{fig:event_study_r1_r5_adjusted}

    \vspace{0.1cm}

    \begin{minipage}{0.96\textwidth}
    \footnotesize
    \textit{Notes:}
    This figure plots event-study treatment-by-event-time coefficients for three indicators of combined 1-star and 5-star review volume using adjusted thresholds. The indicators equal one if a company--day--verification-status observation belongs to the low-volume (0--17 combined 1-star and 5-star reviews), moderate-volume (18--84 combined 1-star and 5-star reviews), or high-volume (85 or more combined 1-star and 5-star reviews) category, respectively, and zero otherwise. The coefficients can therefore be interpreted as changes in the probability that an observation belongs to the corresponding review-volume category. Points represent estimated treatment effects at each event time, and vertical bars show 95\% confidence intervals. The treatment effects are normalized such that the average coefficient over the seven pre-treatment days ($k=-7,\ldots,-1$) equals zero. To impose this normalization, the pre-treatment indicators are parameterized relative to $k=-1$, and the coefficient for $k=-1$ is recovered from the restriction that the seven pre-treatment coefficients sum to zero. The corresponding event-time effects are parameterized using the same pre-treatment normalization. Treatment equals one for unverified reviews and zero for verified reviews. All specifications include company and week fixed effects. Standard errors are clustered at the company level. The vertical dotted line marks the LLM supply shock dates, and the horizontal dotted line denotes zero. The p-value reported below each panel corresponds to the joint test of the pre-treatment treatment-by-event-time coefficients.

    \end{minipage}

\end{figure}

\begin{table}[H]
\centering
\caption{Difference-in-Differences Estimates: Price Reduction Dates}
\label{tab:price_reduction_dates}
\begin{threeparttable}
\begin{adjustbox}{max width=\textwidth}
\begin{tabular}{lcccccc}
\toprule
& \multicolumn{3}{c}{Review rating outcomes} & \multicolumn{3}{c}{Review-volume categories} \\
\cmidrule(lr){2-4} \cmidrule(lr){5-7}
& Avg. rating & Prop. 1-star & Prop. 5-star & \shortstack{Low-volume\\$[0,20)$} &
\shortstack{Moderate-volume\\$[20,100)$} &
\shortstack{High-volume\\100+} \\
\midrule
Treatment
& -0.98827*** & 0.25209*** & -0.21595*** & 0.00136*** & -0.00120*** & -0.00017*** \\
& (0.02503) & (0.00618) & (0.00625) & (0.00015) & (0.00014) & (0.00004) \\

After
& -0.00688 & 0.00200 & -0.00162 & 0.00018*** & -0.00017*** & -0.00001 \\
& (0.00755) & (0.00188) & (0.00245) & (0.00005) & (0.00005) & (0.00001) \\

Treatment $\times$ After
& -0.00477 & -0.00018 & -0.00157 & 0.00005 & -0.00005 & -0.00000 \\
& (0.00797) & (0.00203) & (0.00244) & (0.00004) & (0.00004) & (0.00001) \\

\midrule
Company FE & Yes & Yes & Yes & Yes & Yes & Yes \\
Week FE & Yes & Yes & Yes & Yes & Yes & Yes \\
Mean (After = 0) & 3.784 & 0.259 & 0.633 & 0.999 & 0.001 & 0.000 \\
R-squared & 0.598 & 0.581 & 0.521 & 0.282 & 0.262 & 0.220 \\
Observations & 286,490 & 286,490 & 286,490 & 6,010,256 & 6,010,256 & 6,010,256 \\
\bottomrule
\end{tabular}
\end{adjustbox}
\vspace{0.1cm}

\begin{flushleft}
\footnotesize
\textit{Notes:} Estimates from the baseline Difference-in-Differences specification with company and week fixed effects. Standard errors are reported in parentheses.\\Statistical significance levels: *** $p<0.01$, ** $p<0.05$, * $p<0.1$.
\end{flushleft}
\end{threeparttable}
\end{table}

\begin{table}[H]
\centering
\caption{Difference-in-Differences Estimates: New Model Release Dates}
\label{tab:new_model_release_dates}
\begin{threeparttable}
\begin{adjustbox}{max width=\textwidth}
\begin{tabular}{lcccccc}
\toprule
& \multicolumn{3}{c}{Review rating outcomes} & \multicolumn{3}{c}{Review-volume categories} \\
\cmidrule(lr){2-4} \cmidrule(lr){5-7}
& Avg. rating & Prop. 1-star & Prop. 5-star & \shortstack{Low-volume\\$[0,20)$} &
\shortstack{Moderate-volume\\$[20,100)$} &
\shortstack{High-volume\\100+} \\
\midrule
Treatment
& -1.18741*** & 0.30004*** & -0.26122*** & 0.00125*** & -0.00106*** & -0.00019*** \\
& (0.02618) & (0.00634) & (0.00664) & (0.00014) & (0.00013) & (0.00004) \\

After
& 0.00677 & -0.00193 & 0.00049 & 0.00030*** & -0.00030*** & -0.00001 \\
& (0.00752) & (0.00187) & (0.00242) & (0.00004) & (0.00004) & (0.00001) \\

Treatment $\times$ After
& -0.03173*** & 0.00704*** & -0.00855*** & -0.00011*** & 0.00011*** & -0.00000 \\
& (0.00756) & (0.00193) & (0.00233) & (0.00004) & (0.00004) & (0.00001) \\

\midrule
Company FE & Yes & Yes & Yes & Yes & Yes & Yes \\
Week FE & Yes & Yes & Yes & Yes & Yes & Yes \\
Mean (After = 0) & 3.734 & 0.271 & 0.622 & 0.999 & 0.001 & 0.000 \\
R-squared & 0.613 & 0.594 & 0.536 & 0.297 & 0.271 & 0.250 \\
Observations & 308,237 & 308,237 & 308,237 & 7,200,354 & 7,200,354 & 7,200,354 \\
\bottomrule
\end{tabular}
\end{adjustbox}
\vspace{0.1cm}

\begin{flushleft}
\footnotesize
\textit{Notes:} Estimates from the baseline Difference-in-Differences specification with company and week fixed effects. Standard errors are reported in parentheses.\\Statistical significance levels: *** $p<0.01$, ** $p<0.05$, * $p<0.1$.
\end{flushleft}
\end{threeparttable}
\end{table}

\begin{table}[H]
\centering
\caption{Difference-in-Differences Estimates: Both Price Reduction and New Model Release Dates}
\label{tab:price_reduction_and_new_model_release_dates}
\begin{threeparttable}
\begin{adjustbox}{max width=\textwidth}
\begin{tabular}{lcccccc}
\toprule
& \multicolumn{3}{c}{Review rating outcomes} & \multicolumn{3}{c}{Review-volume categories} \\
\cmidrule(lr){2-4} \cmidrule(lr){5-7}
& Avg. rating & Prop. 1-star & Prop. 5-star & \shortstack{Low-volume\\$[0,20)$} &
\shortstack{Moderate-volume\\$[20,100)$} &
\shortstack{High-volume\\100+} \\
\midrule
Treatment
& -0.82463*** & 0.21122*** & -0.17860*** & 0.00146*** & -0.00129*** & -0.00016*** \\
& (0.03394) & (0.00834) & (0.00858) & (0.00018) & (0.00016) & (0.00005) \\

After
& 0.03083** & -0.00893** & 0.00903** & 0.00013* & -0.00010 & -0.00003 \\
& (0.01397) & (0.00356) & (0.00433) & (0.00007) & (0.00007) & (0.00002) \\

Treatment $\times$ After
& -0.01092 & 0.00197 & -0.00336 & -0.00006 & 0.00004 & 0.00001 \\
& (0.01165) & (0.00296) & (0.00359) & (0.00006) & (0.00006) & (0.00002) \\

\midrule
Company FE & Yes & Yes & Yes & Yes & Yes & Yes \\
Week FE & Yes & Yes & Yes & Yes & Yes & Yes \\
Mean (After = 0) & 3.785 & 0.258 & 0.634 & 0.999 & 0.001 & 0.000 \\
R-squared & 0.634 & 0.616 & 0.559 & 0.329 & 0.307 & 0.301 \\
Observations & 130,283 & 130,283 & 130,283 & 2,905,576 & 2,905,576 & 2,905,576 \\
\bottomrule
\end{tabular}
\end{adjustbox}
\vspace{0.1cm}

\begin{flushleft}
\footnotesize
\textit{Notes:} Estimates from the baseline Difference-in-Differences specification with company and week fixed effects. Standard errors are reported in parentheses.\\Statistical significance levels: *** $p<0.01$, ** $p<0.05$, * $p<0.1$.
\end{flushleft}
\end{threeparttable}
\end{table}

\begin{sidewaystable}[!p]

\centering

\caption{\label{tab:event_study_ratings}
Event-study diagnostics: rating outcomes}

\resizebox{\ifdim\width>\linewidth\linewidth\else\width\fi}{!}{

\fontsize{7}{9}\selectfont

\begin{tabular}[t]{llccccccccc}

\toprule

\multicolumn{2}{c}{}
& \multicolumn{3}{c}{Average rating}
& \multicolumn{3}{c}{Share 1-star}
& \multicolumn{3}{c}{Share 5-star}
\\

\cmidrule(l{3pt}r{3pt}){3-5}
\cmidrule(l{3pt}r{3pt}){6-8}
\cmidrule(l{3pt}r{3pt}){9-11}

Date
& Event type
& Pre p-val
& Post p-val
& Avg. post coef.
& Pre p-val
& Post p-val
& Avg. post coef.
& Pre p-val
& Post p-val
& Avg. post coef.
\\

\midrule

2023-03-01
& Price reduction
& \textbf{0.308}
& \textbf{0.044$^{**}$}
& \textbf{-0.017825}
& 0.374
& 0.215
& 0.001101
& \textbf{0.440}
& \textbf{0.035$^{**}$}
& \textbf{-0.007198}
\\

2023-06-13
& Price reduction
& 0.878
& 0.991
& 0.004475
& 0.792
& 0.986
& -0.001747
& 0.944
& 0.915
& 0.002824
\\

2023-11-06
& Price reduction and new model release
& 0.356
& 0.953
& -0.006253
& 0.535
& 0.971
& 0.001080
& 0.346
& 0.959
& 0.000785
\\

2024-01-24
& Price reduction and new model release
& 0.075$^{*}$
& 0.357
& -0.007144
& 0.065$^{*}$
& 0.411
& -0.000411
& 0.129
& 0.222
& -0.005561
\\

2024-05-13
& New model release
& \textbf{0.278}
& \textbf{0.056$^{*}$}
& \textbf{-0.038776}
& \textbf{0.706}
& \textbf{0.052$^{*}$}
& \textbf{0.009912}
& 0.261
& 0.200
& -0.011298
\\

2024-07-18
& New model release
& \textbf{0.251}
& \textbf{0.029$^{**}$}
& \textbf{-0.012269}
& \textbf{0.549}
& \textbf{0.007$^{***}$}
& \textbf{-0.000351}
& 0.322
& 0.137
& -0.009016
\\

2024-08-09
& Price reduction
& 0.628
& 0.440
& -0.012023
& 0.489
& 0.277
& 0.001733
& 0.733
& 0.513
& -0.003415
\\

2024-09-12
& New model release
& 0.086$^{*}$
& <0.001$^{***}$
& -0.036287
& \textbf{0.291}
& \textbf{0.001$^{***}$}
& \textbf{0.010303}
& 0.084$^{*}$
& <0.001$^{***}$
& -0.005350
\\

2024-10-30
& Price reduction
& 0.268
& 0.242
& 0.018571
& 0.392
& 0.457
& -0.003652
& 0.181
& 0.183
& 0.008047
\\

2024-12-18
& New model release
& 0.808
& 0.184
& -0.036402
& 0.288
& 0.139
& 0.011118
& 0.503
& 0.789
& -0.006115
\\

\bottomrule

\end{tabular}

}

\vspace{0.5em}

\begin{minipage}{0.95\textwidth}

\scriptsize

\textit{Notes:}
The pre-period p-value reports the joint test of the null hypothesis
that the pre-event treatment-effect coefficients are equal to zero.
Under the average pre-treatment normalization, this corresponds to
testing for differential pre-event dynamics between unverified and
verified observations.

The post-period p-value reports the joint test of the null hypothesis
that the post-event treatment-effect coefficients are equal to zero.

The average post coefficient is the mean of the seven estimated
post-event treatment-effect coefficients,
$\hat{\delta}_{1},\ldots,\hat{\delta}_{7}$.

Each row is estimated separately for the indicated event date.
The event-date-specific regressions include company fixed effects
but omit week fixed effects. Standard errors are clustered by company.

Bold entries indicate cases in which the pre-period joint test is
not statistically significant at the 10\% level
(p-value $\geq$ 0.10) and the post-period joint test is statistically
significant at the 10\% level (p-value < 0.10).

Significance levels are denoted by
$^{*}p<0.10$,
$^{**}p<0.05$,
and $^{***}p<0.01$.

\end{minipage}

\end{sidewaystable}

\begin{sidewaystable}[!p]

\centering

\caption{\label{tab:event_study_reviewcounts}
Event-study diagnostics: review-volume outcomes}

\resizebox{\ifdim\width>\linewidth\linewidth\else\width\fi}{!}{

\fontsize{7}{9}\selectfont

\begin{tabular}[t]{llccccccccc}

\toprule

\multicolumn{2}{c}{}
& \multicolumn{3}{c}{Low-volume [0--20)}
& \multicolumn{3}{c}{Moderate-volume [20--100)}
& \multicolumn{3}{c}{High-volume 100+}
\\

\cmidrule(l{3pt}r{3pt}){3-5}
\cmidrule(l{3pt}r{3pt}){6-8}
\cmidrule(l{3pt}r{3pt}){9-11}

Date
& Event type
& Pre p-val
& Post p-val
& Avg. post coef.
& Pre p-val
& Post p-val
& Avg. post coef.
& Pre p-val
& Post p-val
& Avg. post coef.
\\

\midrule

2023-03-01
& Price reduction
& 0.009$^{***}$
& 0.044$^{**}$
& 0.000239
& 0.029$^{**}$
& 0.073$^{*}$
& -0.000203
& 0.362
& 0.648
& -0.000036
\\

2023-06-13
& Price reduction
& <0.001$^{***}$
& <0.001$^{***}$
& -0.000004
& 0.001$^{***}$
& 0.004$^{***}$
& 0.000016
& 0.219
& 0.130
& -0.000013
\\

2023-11-06
& Price reduction and new model release
& 0.027$^{**}$
& 0.041$^{**}$
& -0.000048
& 0.027$^{**}$
& 0.050$^{**}$
& 0.000066
& 0.298
& 0.251
& -0.000018
\\

2024-01-24
& Price reduction and new model release
& 0.002$^{***}$
& 0.085$^{*}$
& -0.000062
& 0.003$^{***}$
& 0.206
& 0.000017
& 0.214
& 0.177
& 0.000045
\\

2024-05-13
& New model release
& <0.001$^{***}$
& 0.002$^{***}$
& -0.000132
& <0.001$^{***}$
& 0.011$^{**}$
& 0.000130
& 0.013$^{**}$
& 0.624
& 0.000002
\\

2024-07-18
& New model release
& 0.002$^{***}$
& 0.006$^{***}$
& 0.000052
& 0.009$^{***}$
& 0.009$^{***}$
& -0.000046
& 0.115
& 0.504
& -0.000005
\\

2024-08-09
& Price reduction
& 0.088$^{*}$
& 0.001$^{***}$
& -0.000024
& \textbf{0.197}
& \textbf{0.003$^{***}$}
& \textbf{-0.000011}
& 0.285
& 0.350
& 0.000036
\\

2024-09-12
& New model release
& <0.001$^{***}$
& 0.006$^{***}$
& 0.000015
& 0.001$^{***}$
& 0.021$^{**}$
& 0.000003
& 0.015$^{**}$
& 0.675
& -0.000017
\\

2024-10-30
& Price reduction
& 0.036$^{**}$
& 0.006$^{***}$
& 0.000035
& 0.066$^{*}$
& 0.014$^{**}$
& -0.000030
& 0.258
& 0.352
& -0.000006
\\

2024-12-18
& New model release
& \textbf{0.263}
& \textbf{<0.001$^{***}$}
& \textbf{-0.000333}
& \textbf{0.292}
& \textbf{0.001$^{***}$}
& \textbf{0.000316}
& 0.985
& 0.685
& 0.000017
\\

\bottomrule

\end{tabular}

}

\vspace{0.5em}

\begin{minipage}{0.95\textwidth}

\scriptsize

\textit{Notes:}
The pre-period p-value reports the joint test of the null hypothesis
that the pre-event treatment-effect coefficients are equal to zero.
Under the average pre-treatment normalization, this corresponds to
testing for differential pre-event dynamics between unverified and
verified observations.

The post-period p-value reports the joint test of the null hypothesis
that the post-event treatment-effect coefficients are equal to zero.

The average post coefficient is the mean of the seven estimated
post-event treatment-effect coefficients,
$\hat{\delta}_{1},\ldots,\hat{\delta}_{7}$.

Each row is estimated separately for the indicated event date.
The event-date-specific regressions include company fixed effects
but omit week fixed effects. Standard errors are clustered by company.

Bold entries indicate cases in which the pre-period joint test is
not statistically significant at the 10\% level
(p-value $\geq$ 0.10) and the post-period joint test is statistically
significant at the 10\% level (p-value < 0.10).

Significance levels are denoted by
$^{*}p<0.10$,
$^{**}p<0.05$,
and $^{***}p<0.01$.

\end{minipage}

\end{sidewaystable}

\clearpage

\setcounter{table}{9}
\renewcommand{\thetable}{S\arabic{table}}

% -------------------------
% Table S10: Quantile 1
% -------------------------

\begin{table}[H]
\centering
\caption{Difference-in-Differences Estimates by Company Size: First (Lowest) Review-Volume Quantile}
\label{tab:q1}

\begin{adjustbox}{max width=\textwidth}
\begin{tabular}{lccccc}
\toprule
& \multicolumn{3}{c}{Review rating outcomes}
& \multicolumn{2}{c}{Review-volume categories} \\
\cmidrule(lr){2-4}
\cmidrule(lr){5-6}
& Avg. rating
& Prop. 1-star
& Prop. 5-star
& \shortstack{Low-volume\\$[0,20)$}
& \shortstack{Moderate-volume\\$[20,100)$} \\
\midrule

Treatment
& -0.9068***
& 0.2333***
& -0.1909***
& -0.000139***
& 0.000139*** \\
& (0.056)
& (0.014)
& (0.013)
& (0.00005)
& (0.00005) \\

After
& -0.0157
& 0.0051
& -0.0016
& -0.000033
& 0.000033 \\
& (0.020)
& (0.005)
& (0.006)
& (0.00006)
& (0.00006) \\

Treatment $\times$ After
& -0.0348*
& 0.0060
& -0.0117**
& 0.000032
& -0.000032 \\
& (0.019)
& (0.005)
& (0.006)
& (0.00008)
& (0.00008) \\

\midrule
Company FE
& Yes & Yes & Yes & Yes & Yes \\

Week FE
& Yes & Yes & Yes & Yes & Yes \\

Mean (After = 0)
& 3.848 & 0.247 & 0.653 & 0.9999 & 0.0001 \\

R-squared
& 0.538 & 0.516 & 0.454 & 0.006 & 0.006 \\

Observations
& 57,144 & 57,144 & 57,144 & 373,264 & 373,264 \\

\bottomrule
\end{tabular}
\end{adjustbox}

\vspace{0.4em}

\begin{minipage}{\textwidth}
\raggedright
\footnotesize
\noindent\textit{Notes:} Estimates from the baseline Difference-in-Differences specification with company and week fixed effects. Standard errors are reported in parentheses. The high-volume 100+ specification could not be estimated because of insufficient variation in the outcome variable.\\Statistical significance levels are denoted as follows: *** $p<0.01$, ** $p<0.05$, and * $p<0.1$.
\end{minipage}

\end{table}

% -------------------------
% Table S2: Quantile 2
% -------------------------

\begin{table}[H]
\centering
\caption{Difference-in-Differences Estimates by Company Size: Second Review-Volume Quantile}
\label{tab:q2}

\begin{adjustbox}{max width=\textwidth}
\begin{tabular}{lccccc}
\toprule
& \multicolumn{3}{c}{Review rating outcomes}
& \multicolumn{2}{c}{Review-volume categories} \\
\cmidrule(lr){2-4}
\cmidrule(lr){5-6}
& Avg. rating
& Prop. 1-star
& Prop. 5-star
& \shortstack{Low-volume\\$[0,20)$}
& \shortstack{Moderate-volume\\$[20,100)$} \\
\midrule

Treatment
& -0.9231***
& 0.2397***
& -0.1917***
& -0.000275***
& 0.000275*** \\
& (0.055)
& (0.014)
& (0.013)
& (0.00008)
& (0.00008) \\

After
& 0.0181
& -0.0053
& 0.0035
& 0.000032
& -0.000032 \\
& (0.016)
& (0.004)
& (0.005)
& (0.00010)
& (0.00010) \\

Treatment $\times$ After
& -0.0134
& 0.0033
& -0.0022
& 0.000138
& -0.000138 \\
& (0.015)
& (0.004)
& (0.005)
& (0.00012)
& (0.00012) \\

\midrule
Company FE
& Yes & Yes & Yes & Yes & Yes \\

Week FE
& Yes & Yes & Yes & Yes & Yes \\

Mean (After = 0)
& 3.901 & 0.232 & 0.664 & 0.9997 & 0.0003 \\

R-squared
& 0.521 & 0.508 & 0.433 & 0.006 & 0.006 \\

Observations
& 79,546 & 79,546 & 79,546 & 379,350 & 379,350 \\

\bottomrule
\end{tabular}
\end{adjustbox}

\vspace{0.4em}

\begin{minipage}{\textwidth}
\raggedright
\footnotesize
\noindent\textit{Notes:} Estimates from the baseline Difference-in-Differences specification with company and week fixed effects. Standard errors are reported in parentheses. The high-volume 100+ specification could not be estimated because of insufficient variation in the outcome variable.\\Statistical significance levels are denoted as follows: *** $p<0.01$, ** $p<0.05$, and * $p<0.1$.
\end{minipage}

\end{table}

% -------------------------
% Table S3: Quantile 3
% -------------------------

\begin{table}[H]
\centering
\caption{Difference-in-Differences Estimates by Company Size: Third Review-Volume Quantile}
\label{tab:q3}
\begin{threeparttable}
\begin{adjustbox}{max width=\textwidth}
\begin{tabular}{lcccccc}
\toprule
& \multicolumn{3}{c}{Review rating outcomes} & \multicolumn{3}{c}{Review-volume categories} \\
\cmidrule(lr){2-4} \cmidrule(lr){5-7}
& Avg. rating & Prop. 1-star & Prop. 5-star & \shortstack{Low-volume\\$[0,20)$} &
\shortstack{Moderate-volume\\$[20,100)$} &
\shortstack{High-volume\\100+} \\
\midrule
Treatment
& -0.9421*** & 0.2370*** & -0.2102*** & -0.000176 & 0.000166 & 0.000010 \\
& (0.049) & (0.012) & (0.012) & (0.00017) & (0.00017) & (0.00002) \\

After
& -0.0022 & -0.0009 & -0.0016 & 0.000194 & -0.000134 & -0.000061** \\
& (0.012) & (0.003) & (0.004) & (0.00018) & (0.00018) & (0.00003) \\

Treatment $\times$ After
& 0.0049 & -0.0013 & 0.0006 & -0.000196 & 0.000155 & 0.000041 \\
& (0.011) & (0.003) & (0.003) & (0.00022) & (0.00022) & (0.00003) \\

\midrule
Company FE & Yes & Yes & Yes & Yes & Yes & Yes \\
Week FE & Yes & Yes & Yes & Yes & Yes & Yes \\
Mean (After = 0) & 4.027 & 0.199 & 0.694 & 0.9993 & 0.0007 & 0.0000 \\
R-squared & 0.495 & 0.479 & 0.409 & 0.014 & 0.014 & 0.005 \\
Observations & 119,181 & 119,181 & 119,181 & 386,050 & 386,050 & 386,050 \\
\bottomrule
\end{tabular}
\end{adjustbox}
\vspace{0.1cm}

\begin{flushleft}
\footnotesize
\textit{Notes:} Estimates from the baseline Difference-in-Differences specification with company and week fixed effects. Standard errors are reported in parentheses.\\Statistical significance levels: *** $p<0.01$, ** $p<0.05$, * $p<0.1$.
\end{flushleft}
\end{threeparttable}
\end{table}

% -------------------------
% Table S4: Quantile 4
% -------------------------

\begin{table}[H]
\centering
\caption{Difference-in-Differences Estimates by Company Size: Fourth (Highest) Review-Volume Quantile}
\label{tab:q4}
\begin{threeparttable}
\begin{adjustbox}{max width=\textwidth}
\begin{tabular}{lcccccc}
\toprule
& \multicolumn{3}{c}{Review rating outcomes} & \multicolumn{3}{c}{Review-volume categories} \\
\cmidrule(lr){2-4} \cmidrule(lr){5-7}
& Avg. rating & Prop. 1-star & Prop. 5-star & \shortstack{Low-volume\\$[0,20)$} &
\shortstack{Moderate-volume\\$[20,100)$} &
\shortstack{High-volume\\100+} \\
\midrule
Treatment
& -1.1463*** & 0.2894*** & -0.2556*** & 0.055033*** & -0.047906*** & -0.007127*** \\
& (0.041) & (0.010) & (0.010) & (0.00548) & (0.00508) & (0.00162) \\

After
& 0.0109 & -0.0018 & 0.0035 & 0.009347*** & -0.008920*** & -0.000427 \\
& (0.007) & (0.002) & (0.002) & (0.00137) & (0.00141) & (0.00043) \\

Treatment $\times$ After
& -0.0217*** & 0.0033 & -0.0077*** & -0.000869 & 0.000911 & -0.000042 \\
& (0.008) & (0.002) & (0.002) & (0.00095) & (0.00096) & (0.00037) \\

\midrule
Company FE & Yes & Yes & Yes & Yes & Yes & Yes \\
Week FE & Yes & Yes & Yes & Yes & Yes & Yes \\
Mean (After = 0) & 4.066 & 0.180 & 0.691 & 0.9488 & 0.0468 & 0.0044 \\
R-squared & 0.446 & 0.422 & 0.393 & 0.252 & 0.228 & 0.217 \\
Observations & 198,571 & 198,571 & 198,571 & 391,438 & 391,438 & 391,438 \\
\bottomrule
\end{tabular}
\end{adjustbox}
\vspace{0.1cm}

\begin{flushleft}
\footnotesize
\textit{Notes:} Estimates from the baseline Difference-in-Differences specification with company and week fixed effects. Standard errors are reported in parentheses.\\Statistical significance levels: *** $p<0.01$, ** $p<0.05$, * $p<0.1$.
\end{flushleft}
\end{threeparttable}
\end{table}

\end{document}